\documentclass[journal,10pt]{IEEEtran}
\usepackage[utf8]{inputenc}
\usepackage{tikz}

\usepackage{booktabs} 
\usepackage{graphicx}
\usepackage{stmaryrd} 
\usepackage{amssymb}
\usepackage{amsmath}
\usepackage{multirow}
\usepackage{colortbl} 
\usepackage{booktabs}
\usepackage{subfig}
\usepackage{comment}
\usepackage{cite}
\usepackage{soul}
\usepackage{hhline}
\usepackage{xspace}
\usepackage{url}
\usepackage{breakurl}
\usepackage{balance}

\newcommand{\AF}[1]{\textcolor{blue}{AF:#1}}
\newcommand{\fakepar}[1]{~\\[-16pt]\subsubsection{#1}}
\newcommand{\censure}[1]{}
\usepackage[export]{adjustbox}

\usepackage{dsfont}

\DeclareMathOperator*{\ARGMAX}{arg\,max}

\title{Deep Learning and Zero-Day Traffic Classification:\\
\huge{Lessons learned from a commercial-grade dataset}}
\author{
\IEEEauthorblockN{Lixuan Yang, Alessandro Finamore, Feng Jun, Dario Rossi\\}
\IEEEauthorblockA{Huawei Technologies, France}
}

\begin{document}
\bstctlcite{IEEEexample:BSTcontrol} 
\pagestyle{plain}
\thispagestyle{empty}

\newcommand{\OURS}{{GradBP}\xspace}

\maketitle


\begin{abstract}
The increasing success of  Machine Learning (ML) and Deep Learning (DL) has recently re-sparked interest towards traffic classification. While supervised techniques provide satisfactory performance when classifying \emph{known} traffic, the detection of \emph{zero-day} (i.e., unknown) traffic is a more challenging task. 

At the same time, zero-day detection, generally tackled with unsupervised techniques such as clustering, received less coverage by the traffic classification literature which focuses more on deriving DL models via supervised techniques. However, the combination of supervised and unsupervised techniques poses challenges not fully covered by the traffic classification literature.

In this paper, we share our experience on a commercial-grade DL traffic classification engine that combines supervised and unsupervised techniques to identify known and zero-day traffic. In particular, we rely on a dataset with \emph{hundreds} of very fine grained application labels, and perform a thorough assessment of two state of the art traffic classifiers in commercial-grade settings. This pushes the boundaries of traffic classifiers evaluation beyond the
\emph{few tens} of classes typically used in the literature.

Our main contribution is the design and evaluation of \OURS, a novel  technique for zero-day applications detection. Based on gradient backpropagation and tailored for DL models, \OURS yields superior performance with respect to state of the art alternatives, in both accuracy and computational cost.
Overall, while ML and DL models are both equally able to provide excellent performance for the classification of known traffic, the non-linear feature extraction process of DL models backbone provides sizable advantages for the detection of unknown classes over classical ML models. 
\end{abstract}


\section{Introduction}\label{sec:intro}

\begin{figure*}
    \centering
    \includegraphics[trim=0 275 0 0,clip,width=\textwidth]{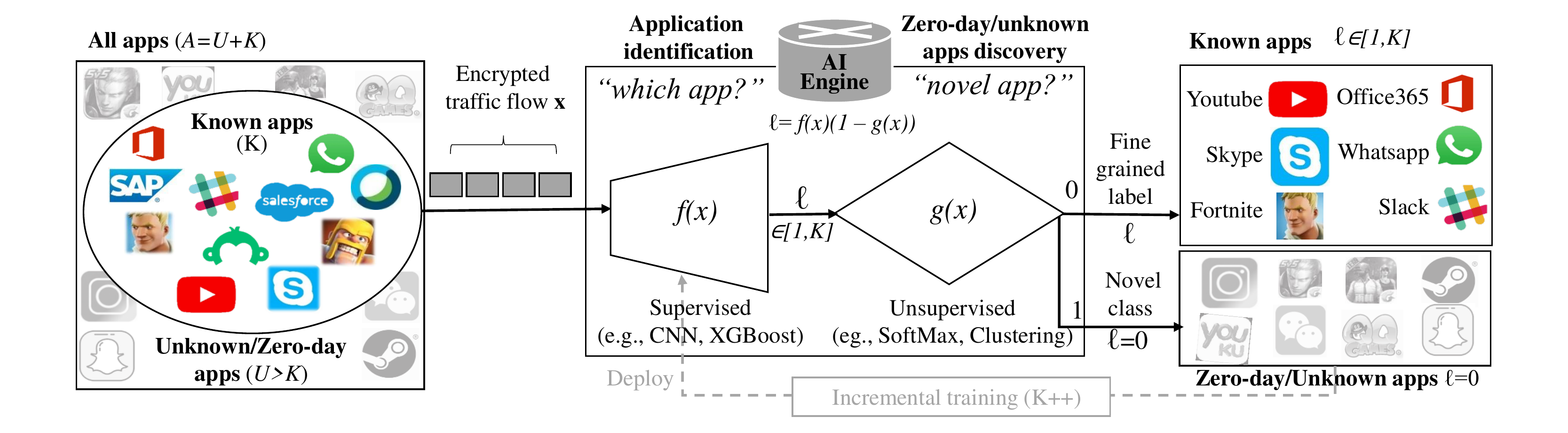}
    \caption{Synoptic of known application identification $f(x)$ and zero-day application detection $g(x)$.}
    \label{fig:system}
\end{figure*}

Internet traffic classification is a well investigated subject.
Research in this area started in the early 2000s, aiming to create novel statistical tools to characterize both broad traffic classes and the specific applications within each class, in order to supplant both light and Deep Packet Inspection (DPI), i.e., the so called port-based and payload-based classification.  
Seminal works such as~\cite{roughan04imc} ignited a \emph{first wave} of approaches~\cite{moore05pam,bernaille06ccr,crotti07ccr,bonfiglio07sigcomm,kim08conext,nguyen08comst} and focused on classic Machine Learning (ML) to classify a relatively small set of applications based on a careful and hand-driven feature engineering process. This first wave culminated with very simple yet effective techniques, referred to as ``early traffic classification''\cite{bernaille06ccr,crotti07ccr} that readily used time series information (e.g., the size and direction of the first few packets in a flow) to take classification decisions.

The tremendous successes of Convolutional Neural Networks (CNN) in the image recognition field~\cite{alexnet12} ignited a \emph{second wave} of traffic classification approaches leveraging  Deep Learning (DL) techniques\cite{wang2015blackhat,taylor2016appscanner,wang2017icoin,wang2017isi,lopez2017access,chen2017bigdata,vu2017ict,aceto2018tma,shapira2019infocom,lotfollahi2020deep}.  DL is becoming particularly appealing in reason of domain-specific CNN hardware accelerators (known as ``tensor processing units'') that started appearing in the last few years, and 
make CNN a viable and appealing option for real-time traffic classification\cite{gallo2020sigcomm}. 
In reason of the tremendous push towards encryption in the Post-Snowden era, this second wave of research is particularly relevant to industry and telco vendors actively looking at deploying  statistical  classification approaches --- as recently pointed out in~\cite{pacheco18comst}, until this point traffic classification remained mostly an academic exercise due to a gap between industry and academia interests, but such scenario is changing.
 
To better understand the reasons behind this gap, Fig.~\ref{fig:system} sketches the fundamental blocks of a classification engine. It is based on two functions: an application identification function $f(x)$, and a zero-day application detection function $g(x)$. First, the goal of $\ell = f(x)$ is to determine, from an input $x$, an application label $\ell\in[1,K]$ among a set of $K$ known classes. Supervised ML/DL techniques fit well this task, and learn $f(x)$ in a process called training.  
However, whereas commercial DPI tools are able to handle \emph{hundreds to thousands} of application classes~\cite{cisco-dpi,spectre-dpi,sandvine-dpi,huawei-dpi}, statistical techniques developed in the academic world consider only a \emph{few tens  of classes} --- this is significantly simpler than commercial needs, hence the gap between industry and academia.

Second, the Internet applications landscape keeps evolving. Thus, commercial engines require a function $g(x)$ to \emph{detect zero-day applications}: $g(x)$ assesses whether the supervised label $\ell=f(x)$  should be rejected, implying that the input $x$ likely belongs to a class never seen  during the training of $f(x)$. In this case $g(x)$ enforces $\ell=0$ to ``signal'' a zero-day detection. Clearly, for this task, unsupervised techniques (such as clustering) are a better fit than supervised ones --- the hard separation of methodologies to perform $f(x)$ and $g(x)$ is introducing frictions, and is recognized as a major blocking point for the deployment of statistical traffic classifiers~\cite{pacheco18comst}. 

In this work, we discuss key aspects on the design of both $f(x)$ and $g(x)$.
First, we share  insights from our experience on creating a classifier $f(x)$ for a commercial-grade dataset comprising tens of millions of flows, and thousands of application labels, significantly larger than what is typically used in the literature. Rather than proposing new techniques, we contrast state of the art ML and DL techniques on exactly the same input (packets size time series for early  classification).


Second, and most important, we  design  \OURS,  a novel zero-day detection technique tailored for DL models, that we thoroughly compare against four state of the art alternatives. In a nutshell, \OURS exploits gradient backpropagation to assess the amount of novelty in the feed-forward inference: the larger the gradient of the first backpropagation step, the smaller the chance that an input $x$ has been seen during training, hence the larger the likelihood that $x$ belongs to a zero-day application. Our analysis shows that \OURS is ($i$) more \emph{accurate} (it detects nearly 80\% of zero-day traffic, compared to less than 11\% for alternatives), ($ii$) \emph{lightweight} (as simple as SoftMax, and over 10$\times$ faster than alternatives), and ($iii$) \emph{incremental}, i.e., it does not require ad-hoc training, or model architecture changes.
Overall, while we find ML and DL techniques being equally capable of accurate identification of known classes, DL models equipped with \OURS have a significant advantage over ML methods for zero-day traffic detection.

In the remainder of this paper we first cover the state of the art on both known application identification $f(x)$ and zero-day traffic detection $g(x)$ (Sec.~\ref{sec:related}). Then, we introduce our commercial-grade dataset (Sec.~\ref{sec:dataset}), and the ML/DL methods selected for $f(x)$ and $g(x)$ (Sec.~\ref{sec:system}). We continue with a separate and thorough evaluation of the identification of known (Sec.~\ref{sec:eval:tc}) and zero-day applications (Sec.~\ref{sec:eval:nd}). Finally, we summarize our findings,  discuss their limits, and introduce open questions for future research avenues (Sec.~\ref{sec:discussion}).

\begin{table*}[!t]
\caption{Related work $\lowercase{f}(x)$: Supervised  identification of known classes}\label{tab:related:tc}
\begin{tabular}{
    @{$\,$}
    l @{$\,\,$} 
    l @{$\,\,$}
    l @{$\,\,$}
    r @{$\,\,$}
    r @{$\,\,$}
    p{7.5em} @{$\,\,$}
    p{20em} @{$\,\,$}}
\hline
\textbf{1st: ML} & 
\textbf{Input} & 
\textbf{Architecture}& 
\textbf{Samples} & 
\textbf{Classes} & 
\textbf{Performance} & 
\textbf{Notes}\\
\hline
2004 \cite{roughan04imc}       & 2 FF  (avg. D, avg S)                 & k-NN vs LDA        & 5.1M     & 4 (7)  & 95\% (91\%) & 100\% adding an extra feature related to $\Delta T$\\ 
2005 \cite{moore05pam}         & PP                                    & 9 DPI heuristics   & 573M     & 10     & 99\% & macro-classes and manual ground truth creation \\
2006 \cite{bernaille06ccr}     & TS$_{5}$ ($\pm$S)                     & K-means            & n.a.     & 10     & $\approx$90\% &\\
2007 \cite{crotti07ccr}        & TS$_{3}$($\pm$S, $\Delta T$)          &  2d gaussian filter    & 30k      & 4      & $\approx$90\% & multi-dimensional statistical fingerprint\\
\multicolumn{6}{c}{}\\

\hline
\textbf{2nd: DL} & 
\textbf{Input} & 
\textbf{Architecture} & 
\textbf{Samples}      & 
\textbf{Classes}  & 
\textbf{Performance} & 
\textbf{Notes}\\
\hline
2015 \cite{wang2015blackhat}        &PP (1000B)                                 & 1D CNN              & 300k  & 58         & 90\% in top-25 & Introduces DL to TC\\
2016 \cite{taylor2016appscanner}    &FF (40)                                    & RF (non-DL)         & 131k  & 110        & 99\%           & Introduces APP classification\\
2017 \cite{wang2017icoin}           &PP (784B=28$\times$28)                     & 2D CNN              & 750k  & 20         & 99\%           & payload as image \\
2017 \cite{wang2017isi}             &PP (784B)                                  & 1D CNN              & 750k  & 12         & $\approx 90\%$  & payload as blob \\
2017 \cite{lopez2017access}         & TS$_{20}$ ($\pm S, \Delta T, p$)            & LSTM + 2D-CNN       & 266k  & 15         & 96\% (82\% w/o $p$) & several models: (CNN+RNN-2a) \\
2017 \cite{chen2017bigdata}         & TS$_{10}$ ($\pm S, \Delta T, p$) + FF (28)  & CNN                 & 22k   & 5+(5 real) & 99\% (88\%) & \\
2017 \cite{vu2017ict}               & FF (22)                                   & GAN (vs DT and RF)  & 682k  & 2        &  all 99\%  & SSH vs non-SSH\\
2018 \cite{aceto2018tma}            & PP, FF, TS                                & MLP, SAE LSTM, CNN  & 138k \censure{77k+44k+27k}& 49& 
80-86\% & 
DL\cite{lopez2017access,wang2015blackhat,wang2017icoin,wang2017isi,lotfollahi2020deep} vs RF~\cite{taylor2016appscanner}\\ 
2019 \cite{shapira2019infocom}      & TS $\rightarrow$ 2D histo ($1500^2$)      & LeNet-5 like        & 21k   & 10       & 99\% &\\
2019 \cite{Rezaei2019} & TS$_6$ ($\pm S$) + PP  ($256\times 6$)      &   1D CNN + LSTM  & unclear  & 80      & 95\% & large scale dataset \\
2020 \cite{lotfollahi2020deep}      & Payload                                   & 1D CNN and SAE      & unclear & 17     & 98\% & dataset is public, but flows are not specified\\
\midrule
\emph{\bf This work}  & \emph{TS}                                   & \emph{1D CNN}      & \emph{$\approx$10M} & \emph{200+635}       & \emph{91\% top-200} & \emph{private commercial-grade dataset}\\
\hline

\multicolumn{7}{@{$\,$}p{61em}@{$\,$}}{\em 
\textbf{Input}: flow features (FF), packet payload (PP), flow duration (D);  time series (TS) of packet size (S), direction ($\pm$),  interarrival ($\Delta T$) and ports (p).}\\
\multicolumn{7}{@{$\,$}p{61em}@{$\,$}}{\em 
\textbf{Architecture}: Random Forest (RF); Multi-layer perceptron (MLP); Stacked Autoencoders (SAE),  Convolutional neural networks (CNN);  Long-short term memory (LSTM); Generative adversarial networks (GAN)}\\
\end{tabular} 
\end{table*}

\section{Related work}\label{sec:related}

\subsection{Identification of known traffic classes $f(x)$}\label{sec:related:tc}

The early 2000's witnessed the first wave of traffic classification, with methodologies aiming to identify fine-grained applications (e.g., YouTube, Skype, Whatsapp) or coarse-grained services (e.g., video streaming,  video call, messaging).
In this section we first summarize the main lessons learned from this period, and then we focus on the second wave where DL-based methods are introduced.
Table~\ref{tab:related:tc} summarizes the relevant literature for both periods.

\fakepar{First wave -- ML} Numerous surveys cover the first wave of traffic classification~\cite{nguyen08comst,boutaba18comprehensive}. This period focused on ``classic'' ML techniques, relying on either flow features (FF) as in~\cite{moore05sigmetrics},  or packet payload features (PP) as in~\cite{moore05pam,bonfiglio07sigcomm}, and it culminated with the adoption of simple, yet effective, time series (TS) features based on properties of the first packets of a flow such as packet size ($S$), direction ($\pm$), and seldom interarrival time ($\Delta T$)~\cite{bernaille06ccr,crotti07ccr}. These lightweight TS approaches are particularly important since they ($i$) operate ``early'' at the beginning of a flow, as opposite to ``post mortem''  techniques which compute FF after a flow ends, and ($ii$) sustain line rate operation with minimal additional  computational costs.
Indeed, whereas payload-based techniques require  large amount of memory (even when processing is done on GPUs~\cite{vasiliadis11ccs}), early TS techniques~\cite{bernaille06ccr,crotti07ccr} have been amenable to line rate classification in excess of 5\,Mfps\cite{santiago12imc} using general purpose CPUs.

\fakepar{Second wave -- DL} 
The second wave of research re-considered all inputs features previously introduces --- from PP~\cite{wang2015blackhat,wang2017icoin,wang2017isi,lotfollahi2020deep}, to FF~\cite{taylor2016appscanner,chen2017bigdata,vu2017ict}, to TS~\cite{lopez2017access,shapira2019infocom}, and hybrid FF+TS ~\cite{chen2017bigdata} --- but still suffered from the same pitfalls experienced during the first wave.

\newcommand{\forgetfootnote}[1]{}
For instance, literature exploiting PP essentially learn text strings, such as the Server Name Indication (SNI)~\cite{rfc6066,Rezaei2019}, i.e., they bind a flow to the hostname advertised (in clear) by the SNI header field. Ultimately, 1d-CNN  approaches\forgetfootnote{Similarly happens for 2d-CNN, that additionally artificially construct ``spatial'' dependencies in the payload that are suitable for images of the physical world, but are less motivated for textual/binary protocols.} leveraging PP result into \emph{a complex mean to do HTTPS protocol dissection, while this can be done trivially via pattern matching.} 
In other words, while DL is an elegant and automated way to statistically learn  SNI ``dictionaries'', the open question is whether CNN inference can be brought to an operational point with  a lower computational footprint than traditional pattern matching techniques. 

Similarly, some works are biased by the use of \emph{port numbers} in their TS ~\cite{lopez2017access,chen2017bigdata}. For instance, authors in~\cite{lopez2017access} show a striking 93\% accuracy for CNN models operating on a single packet input. This ties to the fact that, in academic datasets, HTTP, SSL, and DNS (known to bound to port 80, 443 and 53 respectively) account for over 80\% of flows --- the DL architectures in those works \emph{ultimately exploit the very same port-based information that they are supposed to supplant}.

More generally, the second wave shares the same key weakness already noted in the first wave: the difficulty of cross comparing models in a fair manner. Indeed, as it clearly emerges from  Table.~\ref{tab:related:tc}, every work uses different datasets, with different sample size (from 20k to 750k samples), and with different target classes (from 2 to 100), achieving performance in excess of $>$99\% (under specific conditions), making an apple-to-apple comparison  difficult.

Fortunately,  commendable work such as \cite{aceto2018tma} started appearing, aiming to an independent evaluation of previously published work.
The comparison carried out in~\cite{aceto2018tma} (of \cite{taylor2016appscanner,wang2017icoin,wang2017isi,lopez2017access}, and by mean of datasets different from the ones used by original authors) reveals a different scenario from the one pictured by the original publications: ($i$) the expected performance drops significantly below $<$90\% for any architecture; ($ii$) there is no clear winner, although 1d-CNN models have consistently better results among the candidate approaches; ($iii$)  1d-CNN has a limited gain against shallow Multi Layer Perceptron (MLP) over the same input (+6\%) or against Random Forest (RF) over FF input (+3\%). We underline that such insights are possible only when broadening the evaluation scope beyond the typical race for the 100\%  classification accuracy.


Despite its merits, \cite{aceto2018tma} still partially falls into an apple-vs-orange comparison. 
For instance, the classic RF model inherited from~\cite{taylor2016appscanner} is based on engineered FF, whereas the CNN models are either based on PP~\cite{wang2017icoin,wang2017isi} or packet TS~\cite{lopez2017access}. As such, it is extremely difficult to attribute improvements to either the learning  technique  (ML vs DL) or the model input (FF vs TS). To counter this problem, sharing the same spirit of \cite{aceto2018tma} \emph{we perform an independent evaluation of two state of the art ML/DL techniques, applied to exactly the same input, on a commercial-grade dataset.} By so doing, we reveal problems that only appear at scale, and that are often not properly captured by the academic literature.

\subsection{Detection of zero-day applications $g(x)$}\label{sec:related:nd}

\begin{table*}[!t]
    \centering
\caption{Related work $\lowercase{g}(x)$: Unsupervised zero-day application detection}
    \label{tab:related:nd}
    \begin{tabular}{llllll}
    \toprule
    & {\bf Year [ref]} &  {\bf Technique} & {\bf Applicability} & {\bf $f(x)$ Modification} & {\bf Complexity}  \\

\midrule 
\multirow{4}{*}{\rotatebox{90}{{\bf Input}}}
&2007\cite{Erman07}  &  ~~~Input clustering & ML & None & Not assessed\\
&2015\cite{Zhang15}  & $\checkmark$  Input clustering & ML & None & High (hundreds of clusters for small $K$)\\
&2017 \cite{Liang17} & ~~~Input modification wrt. Temperature scaling & DL & None & High (full backpropagation + inference)\\
&2018\cite{Lee18}    & ~~~Input modification wrt. Mahalanobis & DL & None & High (full backpropagation + inference)\\

\midrule 
\multirow{6}{*}{\rotatebox{90}{{\bf Inner}}}
&2017 \cite{Ge17}& ~~~GAN-OpenMax & ML/DL  & Yes & Medium(Weibull inference)\\
&2018 \cite{Neal2018} &  ~~~Classifier  K+1;& ML/DL &Yes& Low(threshold)\\
&2018 \cite{Hassen18, Aljalbout18, Kamran17, Yang16} & ~~~Clustering loss Functions &ML/DL& Yes & Low(threshold)\\  
&2018 \cite{devries2018learning}& ~~~Confidence learning & ML/DL & Yes & Low(threshold)\\
&2019 \cite{Yoshihashi19} &~~~CNN + AE & ML/DL & None & High(AE inference)\\
&2019 \cite{Zhao19} & ~~~AE & ML/DL & None & High(AE inference)\\
&2020 \cite{Jang20} & ~~~Sigmoid activation & ML/DL & Yes & Medium(Weibull inference)\\
\midrule

\multirow{4}{*}{\rotatebox{90}{{\bf  Output}}}
         &2015 \cite{Rudd15}&~~~Extreme Value Machine (FV)& ML/DL & None & Medium (Weibull inference) \\
         &2015 \cite{Bendale15} & $\checkmark$ OpenMax (AV) & ML/DL & None & Medium (Weibull inference) \\
         &2017 \cite{Hendrycks2017}& $\checkmark$ SoftMax & ML/DL & None& Low (threshold)\\
         &2020 \cite{Zhang20}& $\checkmark$ Clustering (FV)& DL & None & High (clustering) \\
         \hhline{~-----}
         &\it \textbf{This work} & \it~~~Gradient backpropagation &\it DL & \it None & \it Low (backpropagation on last layer)\\

\bottomrule
\multicolumn{6}{l}{\em 
{\bf Technique}: Feature Vector (FV); Activation Vector (AV); Autoencoders (AE),  Convolutional neural networks (CNN); Generative adversarial networks (GAN)}\\
\multicolumn{6}{l}{\em 
{\bf \checkmark}: Technique we compare against in this work}\\
\end{tabular}

\end{table*}


Zero-day applications detection is a network-domain problem arising when either new applications appear, or known applications change their behavior. In both cases a zero-day detection enabled classifier identifies a ``new'' sample as ``unknown'',  and avoids to label it with one of the known classes (in the remainder, we will also refer to an unknown class as a zero-day class).
Zero-day detection is also known as \emph{open-set recognition} in the knowledge discovery domain. In this work, we contribute a new zero-day detection method  (Sec.\ref{sec:system:nd}) with better performance than state of the art alternatives~(Sec~\ref{sec:eval:nd}).
 
We summarize the relevant literature about zero-day detection in Table.~\ref{tab:related:nd} dividing it into three categories based on whether $g(x)$ is performed on the input $x$, at the output $\ell=f(x)$, or at inner-layers of the model. Note that techniques acting at the input or at the output are decoupled from the design of $f(x)$, which is a desirable property as altering existing  models  would make deployment more difficult. Conversely, this is not necessarily true for techniques using models inner-layers.
Moreover, the zero-day detection computational complexity is another important aspect: $g(x)$ is applied on each input, hence it needs to be computed faster than $f(x)$ to avoid slowing down the overall classification process.
In the following we introduce the relevant literature, while deferring a more in-depth discussion to Sec.~\ref{sec:system:nd}.

\fakepar{Input}
A first set of techniques focus on input space properties. This includes seminal works for zero-day detection in the traffic classification context such as~\cite{Erman07}, as well as the current state of the art~\cite{Zhang15} where authors integrate into a classifier a zero-day detection module to continuously update traffic classes knowledge. Shortly, \cite{Erman07,Zhang15} use K-means to cluster input data, and they identify unknown applications by thresholding the distance between an input sample and the clusters centroid. 
These approaches are the state of the art in the network domain, thus they are the reference benchmark. 

A key challenge when using input clustering is the creation of clusters that fit well the different classes. To better control this, some works have recently proposed to apply transformations on the input so to control models output by mean of a Mahalanobis distance based score~\cite{Lee18} or a temperature scaled SoftMax score~\cite{Liang17}. The drawback of those proposals is their additional computational cost beyond\cite{Erman07,Zhang15},  making them less appealing from a practical perspective.

\fakepar{Inner}
At their core, DL methods project input data into a \emph{latent space} where it is easier to separate data based on class labels. A set of works then propose specific ways to alter  this  latent space to ease zero-day detection.  

For instance, \cite{Zhao19} uses AutoEncoders (AE) to transform input data and applies clustering to the transformed input, while \cite{Yoshihashi19}   uses latent representation along with OpenMax~\cite{Bendale15} activation vectors.   Generative Adversarial Networks (GAN) are used in \cite{Ge17,Neal2018} to explore the latent space in order to generate ``unknown classes'' data to train a K+1 classifier for the zero-day class $\ell$$=$0. For instance,  \cite{Ge17}  generates unknown classes by mixing the latent representation of known classes,
while ~\cite{Neal2018}  uses optimization methods to create counterfactual samples that are close to training samples but do not belong to training data. All these methods require specific architectures and extra training, thus they are less appealing.

Other works alter activation~\cite{Jang20} or loss functions~\cite{Hassen18, Aljalbout18, Kamran17, Yang16}. For instance, 
in~\cite{Jang20} authors replace the SoftMax activation with a Sigmoid, and fit a Weibull distribution for each activation output to revise the output activation.
Special clustering loss functions~\cite{Hassen18, Aljalbout18, Kamran17, Yang16} can be used to further constraint points of the same class to be close to each other, so to project unknown classes into sparse regions far from known classes.
All these methods require special DL architectures, and are not incremental. Additionally such architectural modifications can alter the accuracy of the supervised classification task, thus we deem this category inapt for the zero-day detection.


\fakepar{Output}
Classifiers commonly complement the output classification label with a \emph{confidence score}, i.e., the probability that the input $x$ belongs to each of the $K$ target classes. This is typically obtained applying SoftMax to the output layer activation vector. As such, a popular zero-day detection on output data is based on thresholding SoftMax outputs~\cite{Hendrycks2017}, i.e., rejecting a classification if the SoftMax score of the selected class is lower than a threshold. OpenMax~\cite{Bendale15}  
revises SoftMax by adding a special ``synthetic'' unknown class (induced by a Weibull modeling). Alternative approaches include Extreme Value Machine (EVM)~\cite{Rudd15},  or a combination of Principal Component Analysis (PCA) with clustering~\cite{Zhang20} to reduce the ``dimensionality curse''. 
All these approaches are appealing given their limited complexity and do not require to change models.
Our proposed method \OURS fits into this class too.

\section{dataset}\label{sec:dataset}
In this work we use a dataset collected from four Huawei's customer deployments in China.
The dataset does not provide any sensible information about end-users, nor raw pcap collections have been collected.
Each vantage point records per-flow logs where each entry relates to a flow 5-tuple (anonymized ipSrc, ipDst, portSrc, portDst, l4Proto) with aggregate metrics (bytes, packets, mean RTT, etc.) and per-packet information ($\pm$PS and $\Delta$T time series of the first 100 packets). Each log entry is also annotated with application labels provided by a Huawei commercial-grade DPI engine~\cite{cisco-dpi,sandvine-dpi,spectre-dpi,huawei-dpi}. 
Overall, the dataset corresponds to traffic activity across four weeks by tens of thousands network devices.
We used the per-packet TS and the provided applications label to train our ML/DL models, without any specific pre-processing.  

\newcommand{\smslant}[1]{\begin{scriptsize}\textit{#1}\end{scriptsize}}

\fakepar{Collection environments}
The dataset is collected via two Huawei's product line equipments enabling network monitoring and management services for the \emph{Enterprise campus} and \emph{Customer OLT/ONT} market segments in China. We underline that traffic encryption in China is not as pervasive as in the Western world yet, so DPI technologies still offer fine-grained view on traffic. 
\textcolor{black}{In our dataset, HTTPS/TLS  corresponds to 33\% of bytes, and 55\% of flows for TCP traffic, while QUIC is negligible with only 2\% of bytes for less than 1\% of flows.}
This explains the availability of a very large number of labels: the dataset comprises \emph{3,231 application labels that is $30\times$ the largest number of classes considered in the literature}\cite{taylor2016appscanner}. 
As often remarked in the literature, the lack of public datasets is a major limitation: while this dataset is a private Huawei asset, 
we are investigating  the possibility to release a highly anonymized (e.g., shuffled and normalized time series, etc.) and semantically deprived (e.g., no textual labels) version to the community, which we discuss further in Sec.\ref{sec:discussion}.

\begin{table}[t]
    \centering
    \caption{Commercial-grade dataset description}
    
    \label{tab:dataset}
    \footnotesize
    \begin{tabular}{lrrrrrr}
    \toprule

     {\bf Scope$^1$} & {\bf Classes} & \textbf{\%}    &  {\bf Flows} & \textbf{\%} & {\bf Bytes} & \textbf{\%}  \\
    \midrule
     \rowcolor{gray!05}
     &10                & 0.3\% &  3.4M   &  32.6\%   &     5.6\,TB     &  53.0\%   \\ 
    \rowcolor{gray!05}
     &20                & 0.6\% &  4.6M   &  43.6\%   &     7.2\,TB     &  67.1\%   \\
    \rowcolor{gray!05}
$f(x)$ &$K'=$\,50       & 1.5\% &  7.2M   &  68.3\%   &     8.8\,TB     &  82.7\%   \\
    \rowcolor{gray!10}
     &100               & 3.1\% &  8.7M   &  82.9\%   &     9.8\,TB     &  91.9\%   \\ 
    \rowcolor{gray!15}
     &$K''=$\,200       & 6.2\% &  9.9M   &  94.0\%   &     10.3\,TB    &  97.1\%   \\ 
\midrule
    \rowcolor{gray!20}
    $g(x)$  &  250      & 7.7\% &  10.2M  &  95.5\%   &     10.4\,TB    &  98.2\%   \\
 \rowcolor{gray!20}
        &   835         & 25.9\% & 10.4M  &  99.3\%   &     10.6\,TB    &  99.8\%   \\
\midrule
    \rowcolor{gray!30}
noise &    1,000         & 30.9\% & 10.5M  &  99.5\%   &     10.6\,TB    &  99.9\%   \\ 
    \rowcolor{gray!30}
     & 3,231             & 100\%  & 10.5M  &  100\%    &     10.6\,TB    &  100\%    \\ 
\bottomrule
\multicolumn{7}{l}{$^1$Denotes the portion of the dataset which is relevant for the
supervised}\\
\multicolumn{7}{l}{traffic identification $f(x)$ vs unsupervised zero-day detection $g(x)$.}\\
\multicolumn{7}{l}{}\\
\multicolumn{7}{l}{}\\
\end{tabular}

\end{table}

\begin{figure}[t]
    \centering
    \includegraphics[width=\columnwidth]{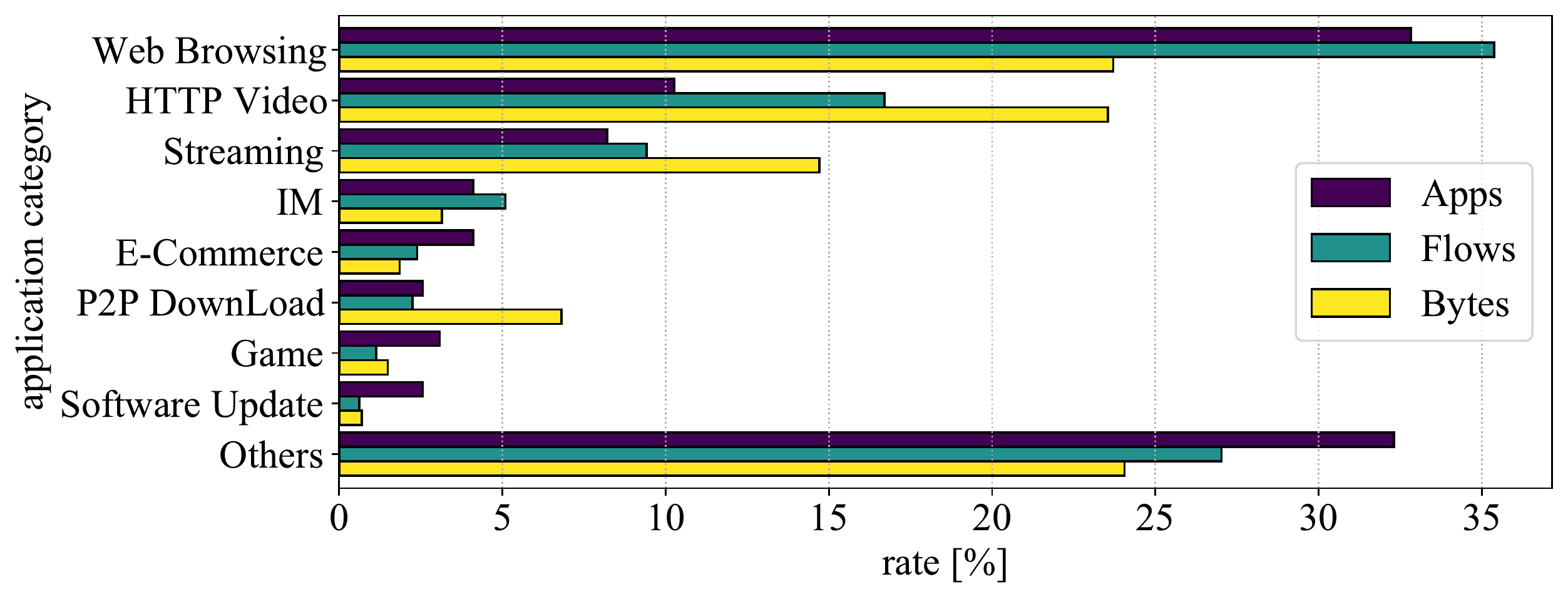}
    \caption{Breakdown of fraction of bytes and flows by category for the top-200 applications.}
    \label{fig:dataset:breakdown}
\end{figure}

\fakepar{Traffic imbalance}
We summarize the dataset properties in Table~\ref{tab:dataset}.
The typical number of classes in academic literature \mbox{$K$$=$$10$--$50$} only covers a tiny portion of the applications catalog (0.3--1.5\%), but captures a sizable portion of flows (32.6--68.3\%) and byte-wise volume (53.0--82.7\%). Yet, even when using $K'$$=$$50$, about 1/3 of flows (1/5 of bytes) are not covered:  a commercial product needs to target about $K''$$=$$200$ to cover roughly 95\% of traffic  (specifically, 94.0\% of flows and 97.1\% of bytes). 
\textcolor{black}{Fig.~\ref{fig:dataset:breakdown} further zooms into the composition of the dataset, reporting a  coarse-grained breakdown of the number of flows, bytes, and labels of the top-200 applications. We can see that applications of the \emph{Web Browsing}, \emph{HTTP Video}, and \emph{Streaming} categories\footnote{The logs offer fine-grained labels, as well as their coarse-grained category.} represent  
51\% of the top-200 apps, and are responsible for 62\% and 61\% of flows and bytes respectively. At the same time, about 32\% of apps, 27\% of flows, and 24\% of bytes are
from \emph{Others}, testifying the great diversity of the data, even when considering only the composition of the top-200 classes.}
In other words, the common scenarios studied in the literature are faraway from business needs: while academic models work well for \mbox{$K$$=$$10$--$50$} classes (accuracy $>$99\%), it is not obvious how to project those results to scenarios with hundreds of classes --- this is at the core of our investigation in Sec.\ref{sec:eval:tc}.

\begin{table}[t]
    \centering
    \caption{Breakdown of most popular ($K$) vs Zero-day TCP/UDP ($U$) applications. }
    \label{tab:dataset:tcp-udp}
    \footnotesize
\begin{tabular}{p{0.7cm}p{1.1cm}p{1.2cm}p{0.9cm}p{0.8cm}p{1.4cm}}
\toprule
 {\bf Proto} & {\bf Known apps $K$} & {\bf Zero-day apps $U$} & {\bf Tot $U+K$} & {\bf Perc $\frac{U}{U+K}$}  & {\bf Openness \mbox{$1-\sqrt{\frac{2K}{2K+U}}$}} \\
    \midrule
TCP  & 162  & 500  & 662  & 75.5\% & 37.3\%\\
UDP  & 38   & 135  & 173  & 78.0\% & 40.0\%\\
\midrule
Total & 200 & 635  & 835  & 76.0\%  & 37.8\%\\
\bottomrule
\end{tabular}
\end{table}


\fakepar{Dataset scope} 
As suggested by the column \emph{Scope} in Table~\ref{tab:dataset}, we are going to focus on the top-200 applications when discussing known application classification $f(x)$.
We then extend the scope to include the top-835 applications (i.e., the applications with at least 100 flows) for zero-day application detection $g(x)$. To better highlight this, Table~\ref{tab:dataset:tcp-udp} further details the split between known ($K$) and unknown ($U$) applications for TCP and UDP traffic. The table additionally reports the  percentage of unknown classes $U/(U+K)$ and the openness $1-\sqrt{2K/(2K+U)}$, which is a metric traditionally used to assess the difficulty of the open-set recognition task. 
 
As mentioned, the top-200 applications (162 TCP and 38 UDP) cover 94\% of flow and 97\% of bytes, which justifies their use for $f(x)$. Indeed, increasing the number of classes beyond 200 minimally affects the  coverage (using $K$$=$$250$ improves the dataset coverage by only 1\%). We introduce DL and ML models for known application identification in Sec.\ref{sec:system:tc}, and evaluate them in Sec.\ref{sec:eval:tc}.  

Conversely, a second portion of 635 applications (500 TCP and 135 UDP) increases the overall coverage to 99.8\% of bytes. We underline that the number of flows per class is too small to properly train and validate a supervised model $f(x)$ including these classes: while the top-200 classes have on average 50,000 labeled flows, these additional 635 classes have only about 750 flows on average. Conversely, this set of applications is well suited to assess zero-day detection  $g(x)$. We stress that this small fraction of roughly  5\% (2\%) of flows (bytes) represents the wide majority (76\%) of the overall labels. We introduce the zero-day discovery techniques in Sec.\ref{sec:system:nd}, and evaluate them in Sec.\ref{sec:eval:nd}.

Finally, a long tail of applications (75\% of the classes, with about 13\% of classes having only one sample) accounts for a tiny fraction of the flows and bytes (about 0.1\%). We consider those as ``noise'' and discard them from the analysis given their limited statistical and practical relevance.


\begin{figure}[t]
  \begin{center}
   \includegraphics[width=0.45\textwidth]{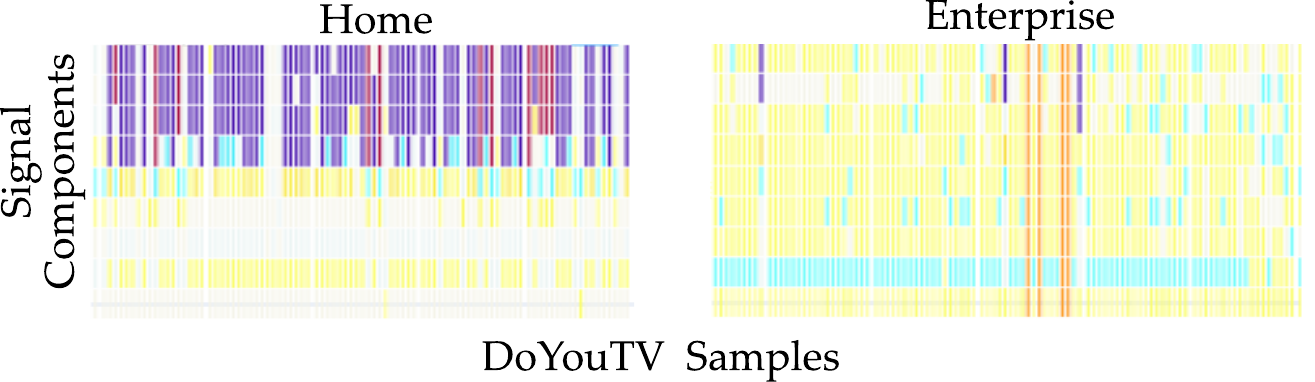}
  \caption{Example of multi-modal behavior for the same application across two environments.}
  \label{fig:dataset}
    \end{center}
\end{figure}

\fakepar{Multi-modality induced by the environment}\label{sec:dataset:multimodal} An interesting phenomenon we observe in our dataset relates to  traffic ``modes'', i.e., applications TS can change due to the network environment. We showcase this for the popular application \emph{DoYouTv} in Fig.\ref{fig:dataset}. The heatmaps depict the top-100 TS across all flows observed from a residential (left) and enterprise campus (right) vantage point. Each column represents one TS, with rows visually encoding the TS values (10 packet sizes) by mean of a color scale. Although not explicitly pointed out in previous literature, multi-modality is an intrinsic effect of access type (encapsulation, maximum segment size, etc.) and the presence of middle boxes (firewalls, NATs, etc.) that may alter packet properties (packet size in our case). This has relevance for both operating and training  traffic classifiers, since those differences need to be incorporated into the training set (e.g., by including samples from all vantage points, or by distributed learning).

\section{Methodology}\label{sec:system}

\subsection{Known application identification $f(x)$}\label{sec:system:tc}
In this section we discuss the models we use in this study. The rationale of our selection is two-fold. First, we seek to avoid yet-another-solution just for the sake of the ``arm race'' against literature. Rather, we consider consolidated modeling techniques, focusing on an apple-vs-apple comparison on a large dataset. Second, we seek to understand models lower bound for a conservative performance assessment. 
As we shall see, this does not limit us from providing key observations. 

\fakepar{Output} 
We focus on fine-grained traffic classification, with models targeting the identification of $K$ classes.
Recalling Table~\ref{tab:related:tc}, the DL classifiers proposed in the literature consider $K$$<$$50$ classes. 
Works considering a larger set of applications either report the accuracy of the top classes (top-25~\cite{wang2015blackhat}) or the dataset used is practically limited to fewer classes (e.g., the top-15 classes represent over 99\% of the traffic in \cite{lopez2017access}). Only a few  works study $K$=50  classes~\cite{aceto2018tma,taylor2016appscanner}.  Conversely, Table~\ref{tab:dataset} shows that we need to consider $K$$=$$200$ classes to cover 95\% of flows (97\% of bytes) in our dataset.
We thus consider $K'$$=$$50$ and $K''$$=$$200$ to both compare against academic state-of-the art~\cite{aceto2018tma}, and cover business needs. We leave as future work an in-depth evaluation of how to identify all 3,231 classes available in the dataset.

\fakepar{Input}  Models input are TS of the first 10 UDP (100 TCP) packets size and direction $\pm$$S$$\in$$\mathbb{Z}$, which values are  normalized into $[0,1]$$\in$$\mathbb{R}$ using the network Maximum Transmission Unit (MTU). Such input TS are easy to collect, and have been consistently found to yield excellent performance across both ML~\cite{bernaille06ccr,crotti07ccr} and DL~\cite{lopez2017access,chen2017bigdata,aceto2018tma} models.

Notice that we could complement $\pm$$S$ input with other packet information (e.g., packets interarrival, TCP header flags, TLS SNI strings).
For instance, the first wave of traffic classification found out that the inter packets time $\Delta T$ could improve classification performance~\cite{bernaille06ccr,crotti07ccr}. Yet, semantically this is prone to error, as 
$\Delta T$ can in practice represent the time between a packet sent and its response (the Round Trip Time -- RTT), which correlates more with the distance between the endpoints than capturing applications behavior. 

Overall, we opt for simple one-dimensional $\pm$$S$ time series as to focus on models lower bound performance (desirable from a scientific standpoint), and to avoid using sensitive/privacy-related info such as TSL SNI strings (desirable from a business standpoint). Differently from previous literature~\cite{aceto2018tma}, we use $\pm$$S$ for both ML and DL models to have an apple-to-apple comparison between modeling techniques.

\fakepar{DL model (1d-CNN)}
As DL model, we use a 1d-CNN given that ($i$) it fits well for time series input, ($ii$) it provides superior performance compared to 2d-CNN~\cite{wang2017icoin,wang2017isi}, and ($iii$) its related literature was independently evaluated~\cite{aceto2018tma}.
\textcolor{black}{We are aware that our choice of a single 1d-CNN architecture is narrow, as more recent designs emerged,
such as  different 2d-CNN inputs~\cite{shapira2019infocom}, sparse-LSTM~\cite{Hua2019} or pointwise correlation~\cite{ma2018shufflenetv2} just to name a few.
While quantitative results reported in this paper pertain to a single architecture, we discuss a broader selection of DL models in Sec.\ref{sec:discussion}.}

\usetikzlibrary{arrows,shadows,positioning}

\tikzset{
  frame/.style={
    text width=6em, text centered,
    minimum height=4em
  },
  line/.style={
    draw, -latex',rounded corners=3mm,
  }
}

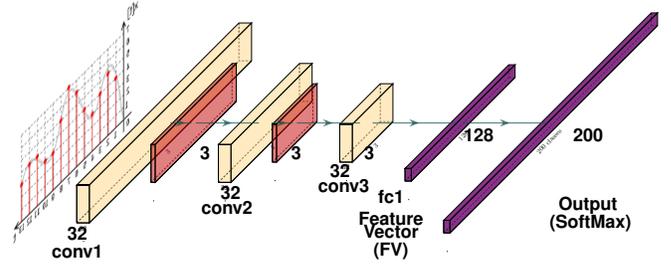
\begin{figure}
\centering
\resizebox{\columnwidth}{!}{%

\begin{tikzpicture}[font=\small\sffamily\bfseries,very thick,node distance = 2cm]

\node[inner sep=0pt] (cnn) at (0,0)
    {\includegraphics[width=\textwidth]{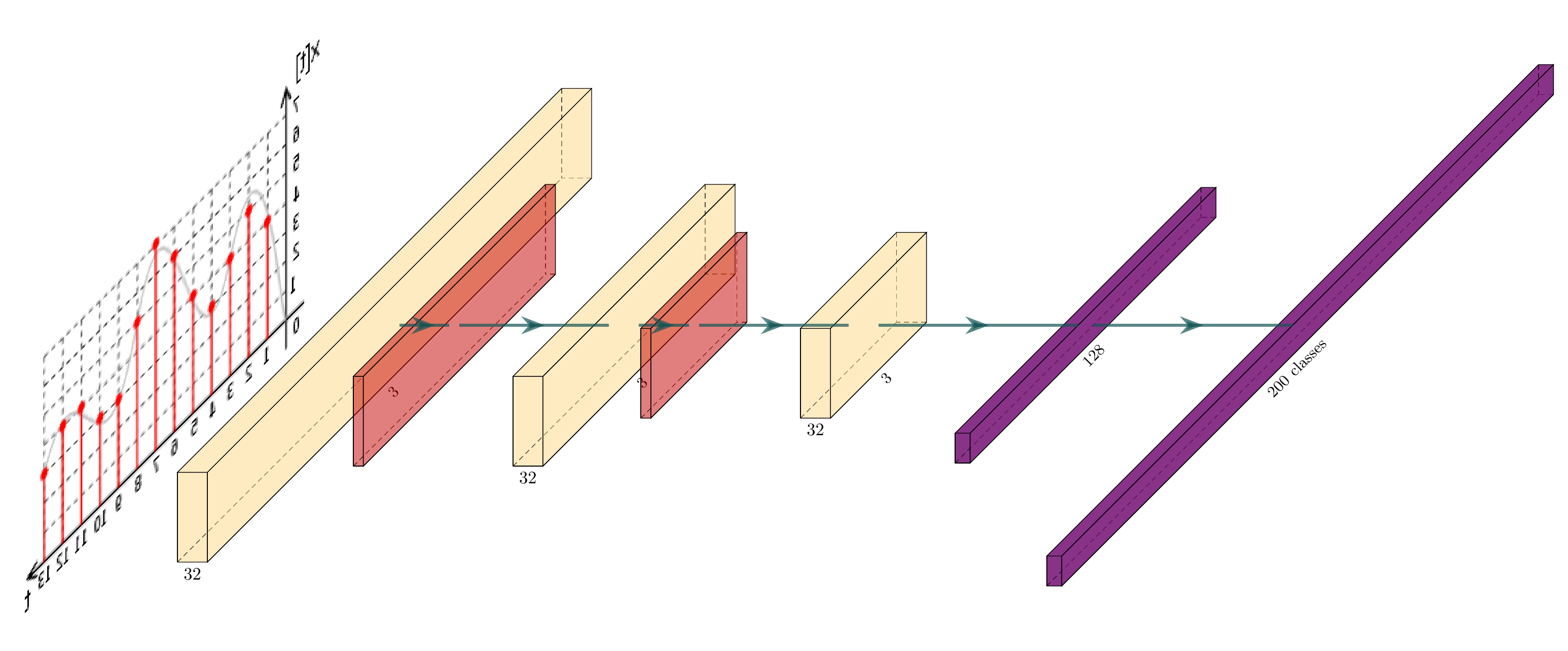}};
\node[below] at (-7,-2.7) {\Large 32};
\node[below] at (-7,-3.2) {\Large conv1};
\node[below] at (-3.5,-0.5) {\Large 3};

\node[below] at (-2.8,-1.6) {\Large 32};
\node[below] at (-2.9,-2) {\Large conv2};
\node[below] at (-1,-0.5) {\Large 3};

\node[below] at (0.2,-1) {\Large 32};
\node[below] at (0.3,-1.4) {\Large conv3};
\node[below] at (1,-0.5) {\Large 3};

\node[frame] at (1.6,-2.5) {\Large fc1 \\ Feature \\ Vector};
\node[below] at (4,-0.) {\Large 128};
\node[frame] at (1.6,-3.5) {\Large (FV)};


\node[frame] at (7,-2.5) {\Large Output \\ (SoftMax)};
\node[below] at (7,-0.) {\Large 200};
\end{tikzpicture}

} 

\caption{Baseline CNN architecture for known application identification $f(x)$  (hyperparameters for TCP traffic portion).
}
\label{fig:ourCNN}
\end{figure}
 
In more detail, to address the different classes and time series length, we model TCP and UDP traffic separately. As per Table~\ref{tab:dataset:tcp-udp}, the top-200 classes include 162 TCP and 38 UDP applications.
Fig.\ref{fig:ourCNN} details the TCP model architecture, but the UDP model has the same structure, with layers size as annotated in parenthesis below.
The input layer feeds a stack of 32 (16) convolutional filters of depth 3 and 1$\times$3 size, with ReLU activations, and max-pooling layers, followed by one fully connected layer of size 128 (64) for a total depth of 4 layers before the final SoftMax classification layer of size 162 (38). We underline that this architecture is rather typical and is adopted with minor differences (hyperparameters selection, and layers size) by many works~\cite{wang2017icoin,wang2017isi,lopez2017access,chen2017bigdata,shapira2019infocom}.
Further notice that we annotated one fully connected layer as FV: this layer plays an important role for zero-day detection (Sec.~\ref{sec:system:nd}).  

\fakepar{ML model (XGBoost)}
As ML model, we use extreme gradient boosting (XGBoost), a tree-based ensemble technique 
widely acknowledged as a ML state of the art classifier in numerous  domains~\cite{friedman2001xgboost,chen2016kdd_xgboost}. 

\fakepar{Models complexity}
We define the ``space'' model complexity as the number of parameters $W_{DL}$ and $W_{ML}$ used by the DL and ML models respectively. While this can be directly measured for the CNN models, we define $W_{ML}$ as the total number of nodes in the trained XGBoost ensemble. Moreover, as a mean to control $W_{ML}$, we fix the number of trees $T$$=$$100$ and cap the individual tree depth $d\in\{2,4,7\}$. 

Conversely, we intentionally avoid time complexity measurements (e.g., training duration, or inference latency) as they directly relate to implementation details  and GPU/TPU hardware acceleration~\cite{gallo2020sigcomm}. This still enables a fair comparison against state of the art, since only space complexity is typically discussed in (or can be easily derived from) the literature.  

\fakepar{Models comparison}
Both ML and the DL models are trained using the same dataset folds. This allows a punctual comparison of classification accuracy and complexity (Sec.\ref{sec:eval:tc}).
Despite the punctual quantification, we underline that the defined models complexity is intended for a qualitative comparison since the individual operations in the ML domain (e.g.,  memory comparison and branching) and DL domain (e.g., tensor products) intrinsically differ. 

\subsection{Zero-day application detection $g(x)$}\label{sec:system:nd}
To identify zero-day applications, the $g(x)$ function needs to reject the classification label for those input $x$ belonging to classes never exposed to $f(x)$ during training (see Fig.\ref{fig:system}).
In other words, $g(x)$ acts as a binary classifier: when $g(x)$$=$$1$ the sample is considered novel and the classification result is rejected by enforcing $\ell$= $f(x)(1-g(x))$=0.  We consider $g(x)$ functions from the literature based on input data, inner/hidden layers telemetry, or the models output.
Additionally, we contribute \OURS, a novel technique which we find superior to alternative methodologies (Sec.~\ref{sec:eval:nd}).

\fakepar{Input based clustering (ML and DL)}
Clustering is a common technique to identify unknown classes using input data~\cite{Erman07,Zhang15}. Clusters are formed using training samples of the known applications. Comparing the distance between an input $x$ and the clusters centroid with a predefined threshold allows to accept/reject the $f(x)$ label. Specifically, the reject criterion is:
\begin{equation}
g^{IN}(x) = \mathds{1} \left( \min_{c\in[1,C]}  d(x,c) > \epsilon_{IN} \right) 
\end{equation}
\noindent where $d(x,c)$ is a distance  metric (Euclidean, Manhattan, etc.), and $\epsilon_{IN}$ is an arbitrary threshold. 
Intuitively, a small $\epsilon_{IN}$ increases the chance to reject a correctly classified sample of a known application (false negative), whereas setting a too large threshold leads to unknown applications going undetected (false positive). 
The reference state of the art technique is~\cite{Zhang15}, which uses K-means to identify a large number of clusters $C$$\gg$$K$ so to fit the input space. Clearly, the number of clusters $C$ trades off computational complexity (as a point needs to be compared against $C$ centroids) and accuracy (as the larger is $C$, the better the coverage of the input space).

\fakepar{Inner layers based clustering (DL only)}
When using a DL model, clustering can be applied also to FVs, i.e., the output generated by one of the hidden layers when processing an input sample (see Fig.\ref{fig:ourCNN}). 
Essentially, the training process of DL models performs non-linear transformations that improve its discrimination power. The output of the FV layer is a \emph{high-dimensional latent space} where the projected input samples are more easily separable with respect to their class. It is possible to induce clustering in the latent space by either using a clustering loss function~\cite{Hassen18, Aljalbout18, Kamran17, Yang16}, or by regularizing the space (e.g., by normalizing the layer into a hyper-sphere). The $C$ clusters are constructed at training time applying K-means to $FV(x)$. The rejection criterion is: 
\begin{equation}
g^{FV}(x) = \mathds{1}\Big(
\min_{c\in[1,C]}   d(FV(x),c) >\epsilon_{FV} \Big)
\end{equation}
\noindent where, as previously,  $d(\cdot)$ and $\epsilon_{FV}$ are a distance metric and an arbitrary threshold respectively. This technique is standard in the DL domain, and can be considered as the state of the art for DL-based clustering\cite{Zhang20}.

\fakepar{Output based rejection (ML and DL)}
Output-based techniques leverage additional information from the ML/DL model output, such as the SoftMax\cite{Hendrycks2017} or OpenMax\cite{Bendale15} probabilities of each class. 
Denoting with $v(x)$  the activation vectors of an input sample $x$, and with $v_i(x)$ the i-th component of the vector, then the SoftMax value for class $k$ given input $x$ is: 
\begin{equation}
P(y=k|x) = \frac{e^{v_k(x)}}{\sum_{i=1}^K e^{v_i(x)} }
\label{eq:softmax}
\end{equation}
\noindent Denoting further with  $c  = \ARGMAX_{k} P(y=k|x)$ the most likely class  selected by the supervised model, then the SoftMax rejection criterion is:
\begin{equation}
\label{eq:proba-softmax}
g^{SM}(x) = \mathds{1}\Big(  P(y=c|x) < \epsilon_{SM}  \Big)
\end{equation}
\noindent that is, the SoftMax  of the output class $c=f(x)$ is required to be larger than an arbitrary threshold $\epsilon_{SM}$. This technique should be considered a na\"ive baseline since supervised models are knowingly \emph{overconfident} --- the class confidence generated by SoftMax can be high despite the classification is wrong, which stems from the nature of SoftMax to saturate~\cite{ulmer2021}.

Part of the problem is rooted in the fact that the SoftMax function normalizes only considering the space of known applications. To solve this issue, OpenMax\cite{Bendale15} introduces an extra ``synthetic zero label'' and re-normalizes the activation vector before computing distances. In more detail, OpenMax introduces
a weight vector $\omega_b$ that
captures the distribution of the activation vectors $v(x)$ of the $K$ known classes by fitting a Weibull distribution. The vector $\omega_b$ includes also an
extra synthetic activation value $\hat{v}_0(x)$, namely the \emph{novel class $\ell=0$}. A new activation vector $\hat{v}(x)$ is then derived as follows:
\begin{equation}
\hat{v}(x) =v(x)\circ w_b(x) 
\end{equation}
\begin{equation}
\hat{v}_0(x)=\sum_{k=1}^{K} v_k(x)(1-w^{k}_{b}(x))
\label{eq:openmax}
\end{equation}
\noindent OpenMax probabilities are then derived with the re-normalized activation vector, including for the novel class $c$=$0$
\begin{equation}
\label{eq:proba-openmax}
P'(y=k|x) = \frac{e^{\hat{v}_k(x)}}{\sum_{i=0}^K e^{\hat{v}_i(x)} }
\end{equation}

Notice also how~(\ref{eq:proba-openmax}) normalizes starting from $i$=$0$ to reflect the extra synthetic activation, while~(\ref{eq:proba-softmax}) normalizes starting from $i$=$1$. Denoting with $c'=\ARGMAX_{k} P'(y=k|x)$ the class with the largest OpenMax values, the rejection criterion then becomes:
\begin{equation}
g^{OM}(x) = \mathds{1}\Big( c'= 0   \vee P'(y=c'|x) < \epsilon_{OM}  \Big)
\end{equation}
\noindent which rejects when either the novel class $c=0$ has the largest value, or when the normalized OpenMax value for the most likely class $c\in[1,K]$ is smaller  than a threshold $\epsilon_{OM}$. 


\fakepar{ \OURS -- Gradient based rejection (DL only)}
Finally, we introduce \OURS, our novel zero-day detection method.
The idea is to perform a ``shadow'' training step: ($i$) evaluate the output label $\ell=f(x)$; ($ii$) treat the label $\ell$ as the groundtruth and compute the magnitude of the first backpropagation step $\delta^{L-1}$ but without actually altering the model weights; ($iii$) use $\delta^{L-1}$ to assess if the input is of a known or unknown application. More precisely, let us consider 
\begin{equation}
    \delta^L = \nabla_a \mathcal{L} \odot \sigma'(z^L)
    \label{eq:gradient}
\end{equation}

\noindent where $\nabla_a \mathcal{L}$ is the partial derivative of the model loss function $\mathcal{L}$ with respect to the activation (i.e., the magnitude of model weights update) and $\sigma'(z^L)$ is the activation vector $v(x)$ at the model layer $L$. We limit the backpropagation to the output of the convolutional layers (i.e., the output of the \emph{model backbone}), which already captures the majority of the information the model uses to represent the input. This corresponds to
\begin{equation}
    \delta^{L-1} = ({W^{(L-1)}}^T \delta^{L}) \odot \sigma'(z^{L-1})
    \label{eq:gradient2}
\end{equation}
\noindent where ${W^{(L-1)}}^T$ is the transpose of the weight matrix of the $(L-1)$-th layer. Intuitively, the larger the $\delta^{L-1}$, the more likely the input relates to an unknown class. Based on this intuition, we conceive a simple family of gradient-based rejection criteria:
\begin{equation}  
  g^{GR}_{n}(x) =  \mathds{1}\Big( \| \delta^{L-1}\|_n > \epsilon_{GR} \Big)
\end{equation}  
\noindent where the norm $\|\cdot\|_n$ and $\epsilon_{GR}$ are free hyperparameters.
In particular, for  L1 (i.e.,  max gradient) and L2 norms (i.e., the square root of the squared gradient sum) we have:
\begin{equation}  
  g^{GR}_{1}(x) =  \mathds{1}\Big( \textrm{max}_i \delta_i > \epsilon_{GR} \Big)
\end{equation}  
\begin{equation}  
  g^{GR}_{2}(x) =  \mathds{1}\left( \sqrt{\sum_i \delta_i^2} > \epsilon_{GR} \right)
\end{equation}  

\noindent Clearly,  backpropagation is an essential tool, and gradients have been used in many aspects of DL, from training (e.g., to speed up 
convergence\cite{wen2019} possibly in distributed settings\cite{ yang2020ijcai})  to extract  side-channel information (e.g., to gather information about clients participating into a  federated learning cohort~\cite{Nasr2019}). However, we are not
aware of such use in neither zero-day applications detection, nor in open-set recognition.

\section{Identification of Known Applications}\label{sec:eval:tc}



\begin{figure}
    \centering
    \includegraphics[width=\columnwidth]{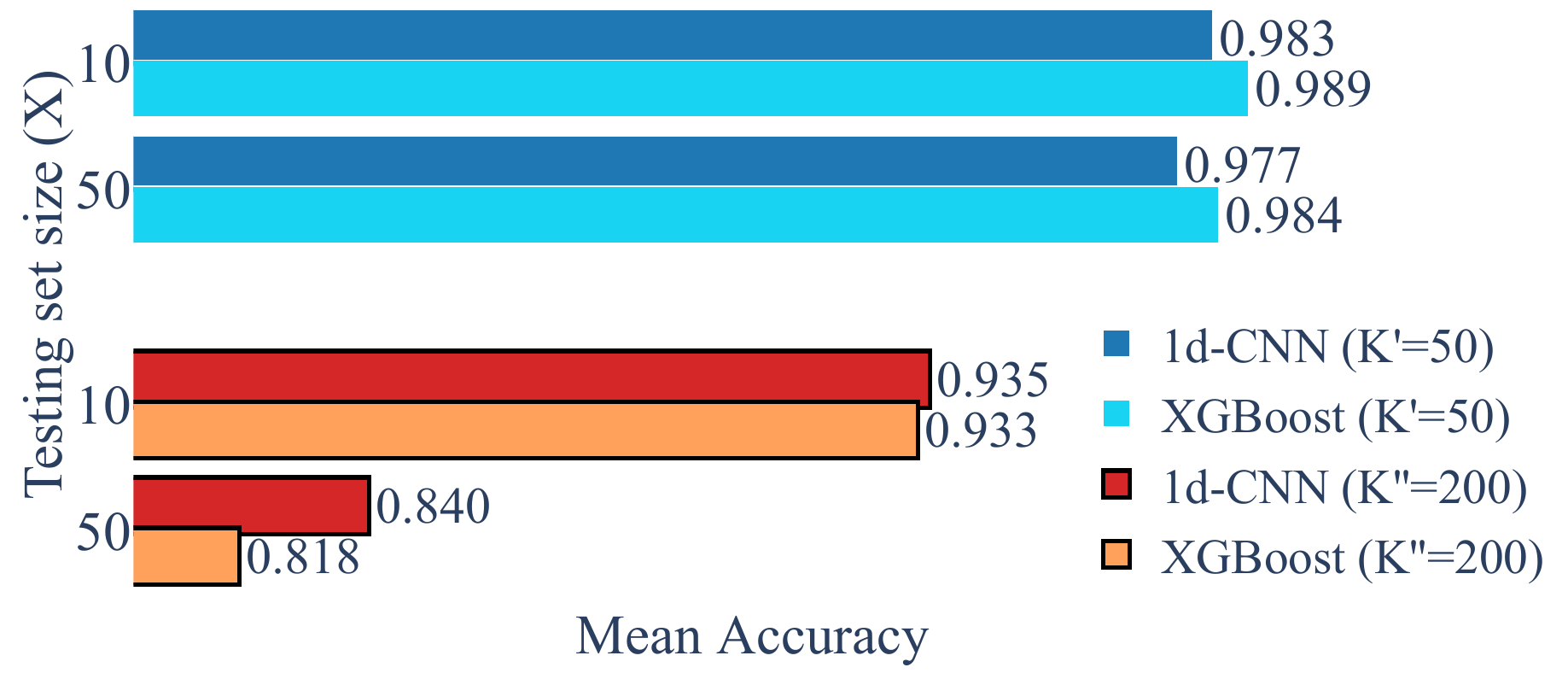}
    \caption{\textit{Class cardinality bias}: When training the same ML/DL model for a large number of classes $K''$$\gg$$K'$  accuracy drops  also for the top $X$$\le$$K$ classes, which is unnoticed in literature.}
    \label{fig:trainKtestX}
\end{figure}


Models in literature are trained to identify $K'$$\le$$ 50$ classes, and evaluated using a test set containing $X$$\in$$\{10,50\}$ classes. In this section, we extend this approach towards a more business-driven scenario 
with $K''$$\le$$200$ classes and $X$$\le$$200$. 
We train and evaluate TCP (162 applications) and UDP classifiers (38 applications) separately to better address their traffic diversity, but we  also report their combined performance (top-200 applications).

The use of a large number of classes $K$, that we consider here as a simple proxy of   
class \emph{variety} (which includes class imbalance, multi-modality, inter-class similarity, etc.), allows us to better capture macroscopic trends that we believe were not well represented in previous literature. In particular, we start by illustrating how restricting the analysis to the \mbox{top-K} classes can bias the classification accuracy (Sec.\ref{sec:accuracy}).
We continue by digging into classification confounding factors and labels semantic  (Sec.\ref{sec:deeper}). Finally, we conclude by discussing models complexity (Sec.\ref{sec:complexity}).

\subsection{High-level view}\label{sec:accuracy}

\begin{figure*}[t]
    \centering
  \subfloat[]{\includegraphics[width=0.5\textwidth]{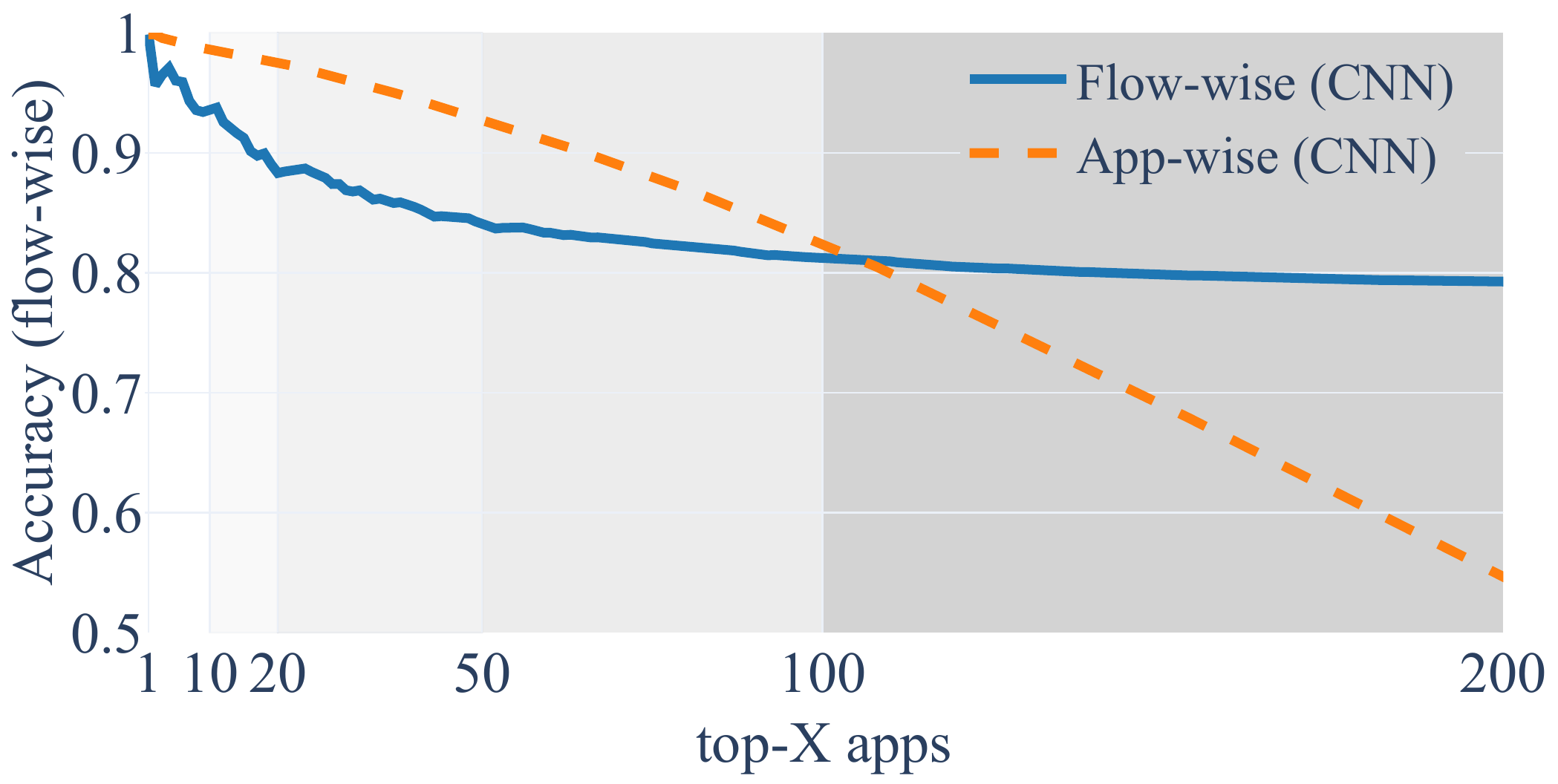}} 
    \subfloat[]{\includegraphics[width=0.5\textwidth]{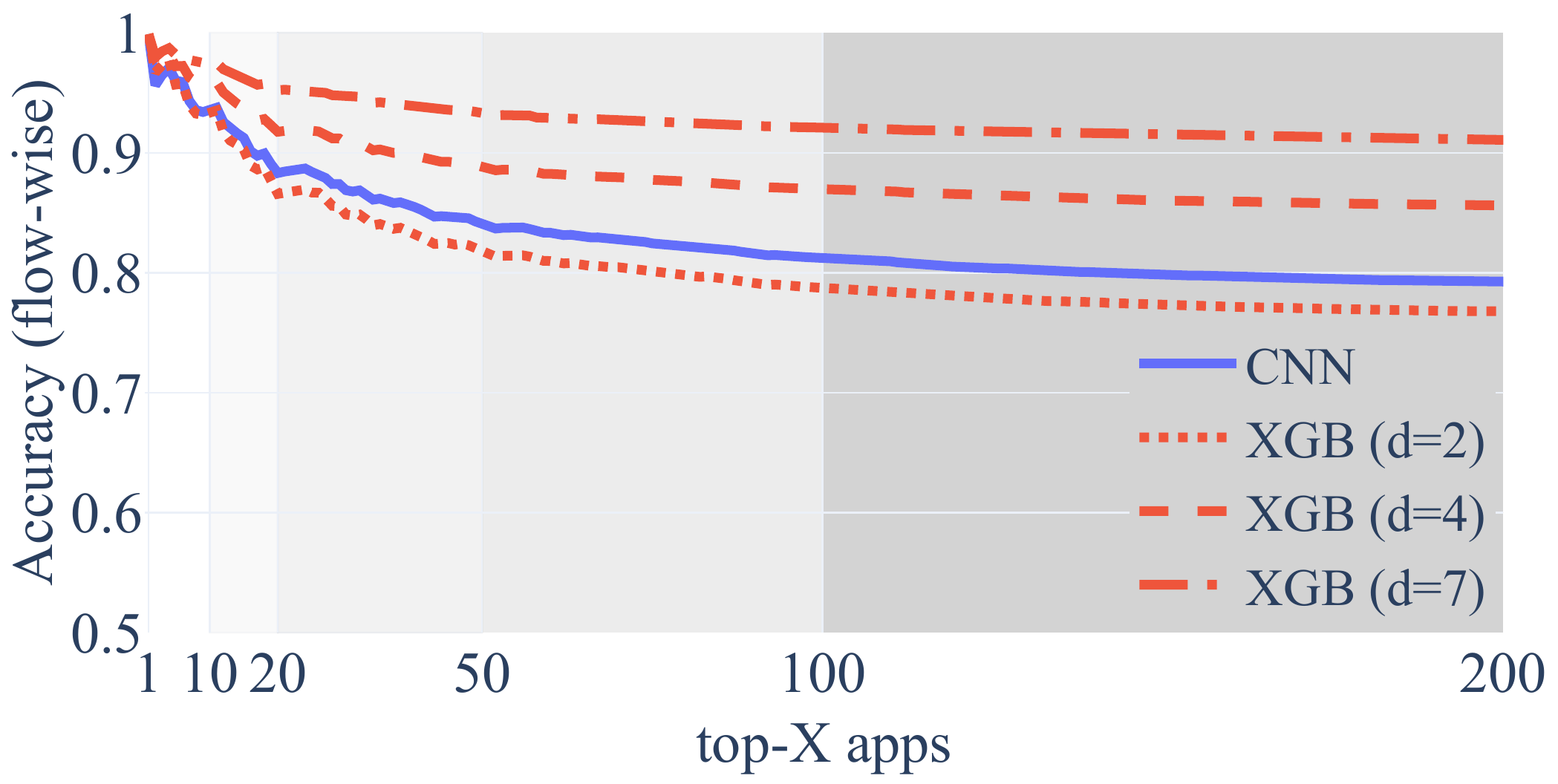}} 
   \caption{\emph{(a) Mean accuracy  bias.} Average per-flow (affected by class imbalance) or per-application accuracy (each application counted equally). Notice that the per-application accuracy overestimates performance  in the small class regime $K$$<$$100$, and vice-versa happens for $K$$>$$100$. \emph{(b) Model bias}. By tuning model hyperparameters, it is easy to obtain operational points 
    that show superiority of one class of approaches (purposely biased to ML in this example).}
    \label{fig:ml-vs-dl}  \label{fig:accuracy}
\end{figure*}
\begin{figure*}
    \centering
    \includegraphics[width=0.22\textwidth]{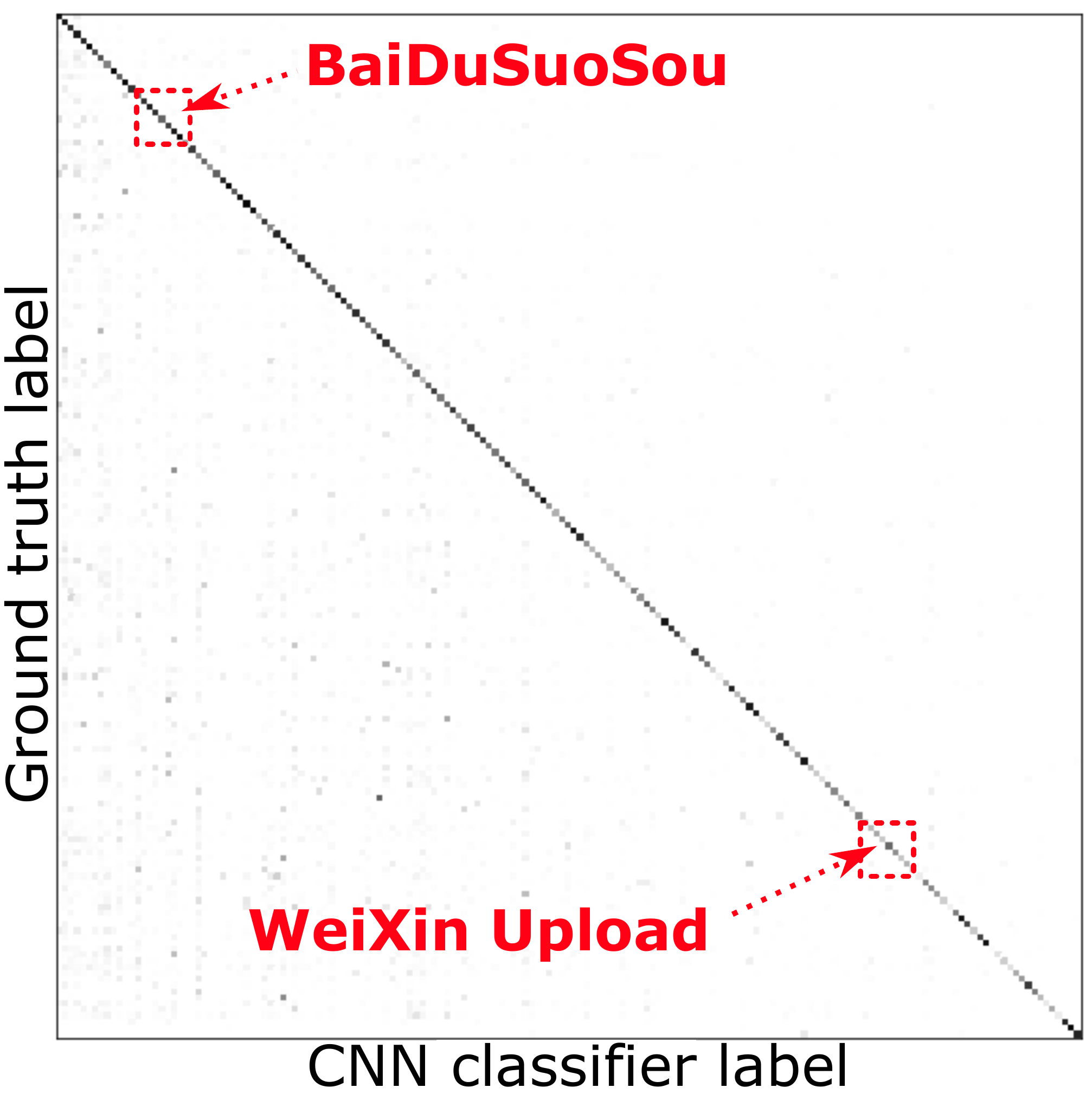}
    \includegraphics[width=0.38\textwidth]{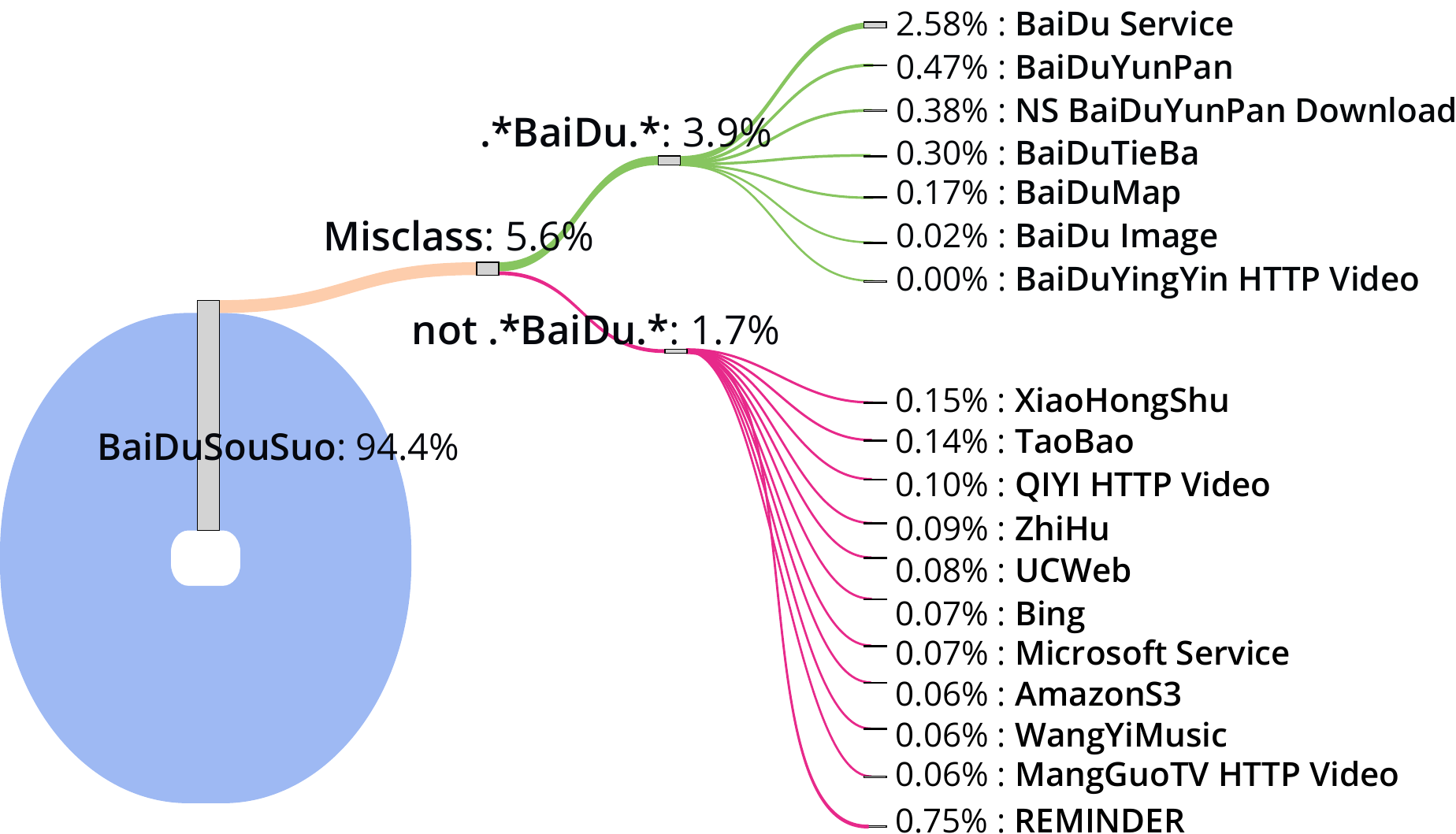}
    \includegraphics[width=0.38\textwidth]{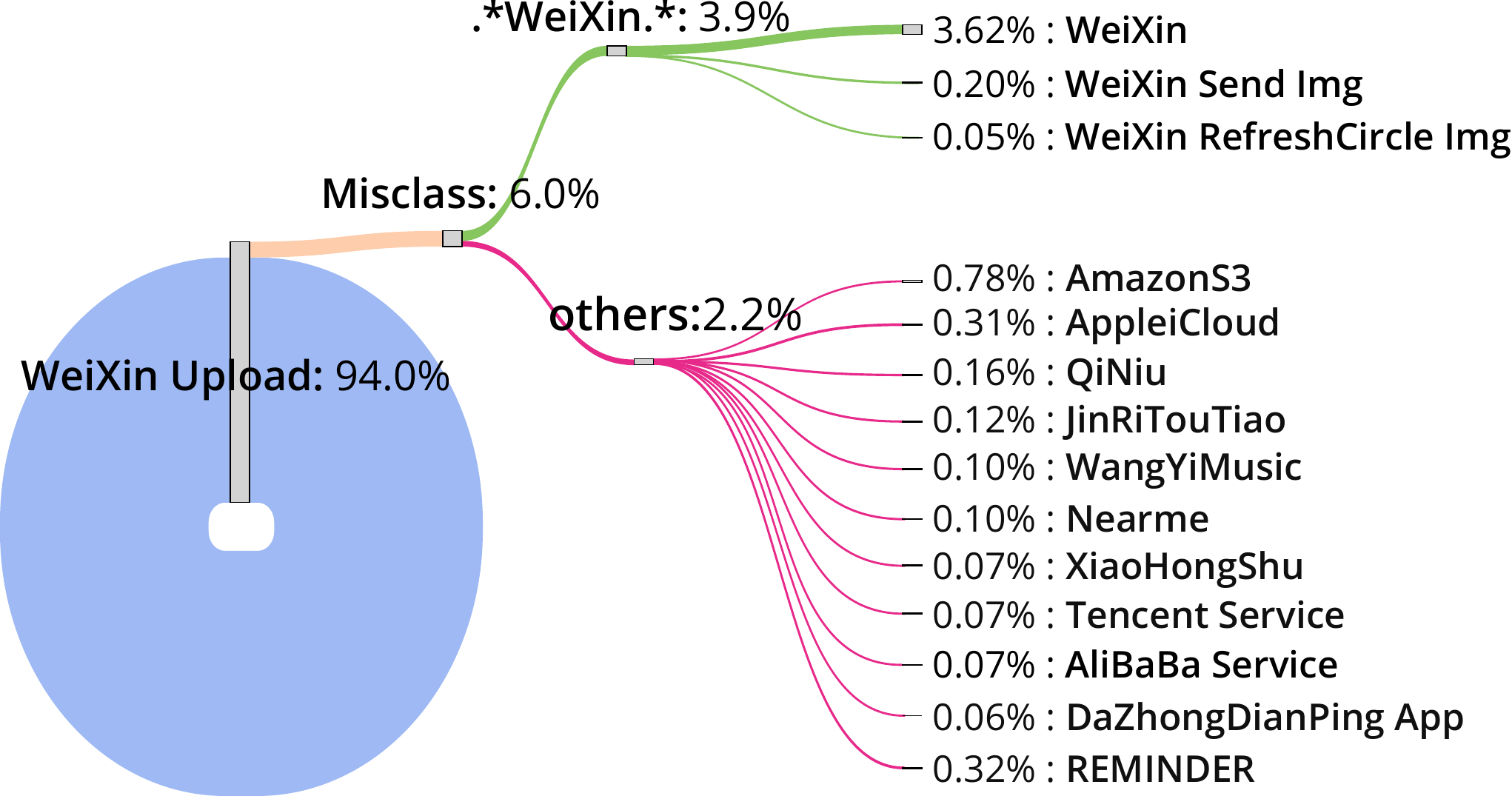}
    \caption{Investigating classification performance for TCP traffic: (left) classification confusion matrix, with applications sorted by volume of bytes --- performance is very good, but application with less traffic can suffer from lower accuracy; (center) BaiDuSouSuo classification breakdown --- misclassifications mostly relate to alternative labeling of traffic of the same application; (right) WeiXin Upload --- a large portion of misclassifications are due to other behaviors of the same application.
    }
    \label{fig:deeper}
\end{figure*}

\fakepar{Class cardinality bias}
We start by comparing ML and DL models trained so that when $X$$=$$K'$$=$$50$ their flow-level accuracy matches what reported in the literature for the top-$X$$=$$50$ classes ($\approx$99\%). Fig.\ref{fig:trainKtestX} shows the models mean classification accuracy when using $X$$\in$$\{10,50\}$. We can see that models performance is stable when training with 50 classes. Conversely, when increasing 4$\times$ the number of classes ($K''$$=$$200$), the accuracy notably degrades for top-10 classes (even further for the top-50).
\emph{In other words, focusing on a small number of classes (as common in literature) makes the problem trivial to solve (even without DL).} 


\fakepar{Mean accuracy bias} To further elaborate, Fig.\ref{fig:accuracy} shows the evolution of the mean accuracy for a model with $K''$$=$$200$ classes (combining TCP and UDP models). We consider the top-X classes when ranked by their popularity (classes with larger number of flows come first), and performance (best classified class comes first). The picture shows that
($i$) despite classification accuracy worsen as $X$ gets closer to $K''$ (dashed line),  ($ii$) the overall number of correctly classified flows remains satisfactory in reason of the application popularity skew (solid line), and ($iii$) this effect appears only for a number of classes larger than the one typically used in the academic literature.
\emph{Otherwise stated, focusing on the average accuracy of a limited number of classes $K'$$\ll$$K''$ hides phenomena typical in commercial settings.}

\fakepar{DL bias}
We argue that the second wave of traffic classification published until this moment may have exaggerated the quality of DL methodologies, 
as already observed in computer science~\cite{troublingtrends}, and other fields\cite{aiadvancesarenotreal}. We exemplify this in Fig.\ref{fig:ml-vs-dl} showing that XGBoost models performance can (significantly) surpass CNN models performance by simply tuning the maximum tree depth --- although we could have, deliberately or unwittingly, adopted the opposite viewpoint. 
\emph{Thus, as far as classification of known traffic is concerned, we find that ML and DL models are equally well fitted to support large-scale commercial scenarios}.

\subsection{Deeper insights}\label{sec:deeper}   

We next dive into labels semantic to dissect misclassification causes. Again, we consider $K''$=$200$ classes by combining TCP and UDP models. Fig.\ref{fig:deeper}-(left) shows the classification confusion matrix as a heatmap, with applications sorted alphabetically (labels not shown for readability): the sharp diagonal indicates that flows are classified with the expected label.

To further drill down, we pick a representative class for each side of the diagonal, \emph{BaiDuSuoSou} (the BaiDu search service) and \emph{WeiXin Upload} (the file upload service of the WeiXin messaging application), and we dissect their classification results by mean of a sankey diagram. For \emph{BaiDuSuoSou},  Fig.\ref{fig:deeper}-(center) shows that 5.6\% of flows are misclassified, but 3.9\% of flows are labeled as \emph{BaiDu}-related services (e.g., searches for image, maps, or social) and overall only 1.7\% of flows are completely misclassified. The same considerations hold for \emph{WeiXin Upload} where only 2.2\% of misclassifications are imputed to non \emph{WeiXin} related classes, and other popular applications (not shown). \emph{In other words, a non negligible part of  misclassifications are for neighboring services of the same application provider}.

On the one hand this phenomenon ties to the very fine-grained nature of the groundtruth labels in our dataset; on the other hand it relates to the existence of 
different ``modes'' of the same application. Intuitively, the $\pm S$ time series capture application-level signaling at the beginning of a flow. Different applications might share common patterns when being part of the same ``umbrella'' of services and relying on the same codebase and libraries (as in the case of \emph{Baidu}), leading to ``soft misclassifications''. \emph{Overall, this suggests that fine-grained applications identification is possible, with high accuracy, even for a larger sets of known applications than what was considered in the literature}.

This holds particularly true for applications where sufficient number of samples are provided to the training process: notice indeed the degradation of flow-wise accuracy for less popular applications in the top-200, as early shown in Fig.\ref{fig:accuracy}-(b), confirming that class imbalance in the real traffic helps maintaining satisfactory 
accuracy even in commercial-grade cases. At the same time, recall that labels are scarce beyond the top-200  applications (Sec.~\ref{sec:dataset}).
Thus, training $f(x)$ with a larger number of classes calls for experimentation with few-shot learning~\cite{wang2020generalizing}, and ensemble-based methodologies~\cite{distillation-NIPS15}, which we discuss further in Sec.~\ref{sec:discussion}.


\subsection{Complexity}\label{sec:complexity}
In our view, traffic classification literature tends to underestimate models complexity.
As from Sec.\ref{sec:system:tc}, for DL models the space complexity is the number of model parameters $W_{DL}$, while for ML models we use the number of nodes in the tree ensemble $W_{ML}$. We reiterate that space complexity does neither directly translate into computational complexity (as this depends on the ML/DL architecture, and the specific operations executed at inference time), nor energy expenditures (as this depends on the available hardware). Yet, space complexity allows to abstract from specific implementations (i.e., software frameworks, hardware acceleration, system design choices), enabling a  \emph{qualitative} comparison between models. In particular, as models have disparate capabilities, we compare not only the absolute models size $W$, but especially the model size normalized over the number of output classes $W/K$.

\begin{figure}
\centering
\includegraphics[width=0.47\textwidth]{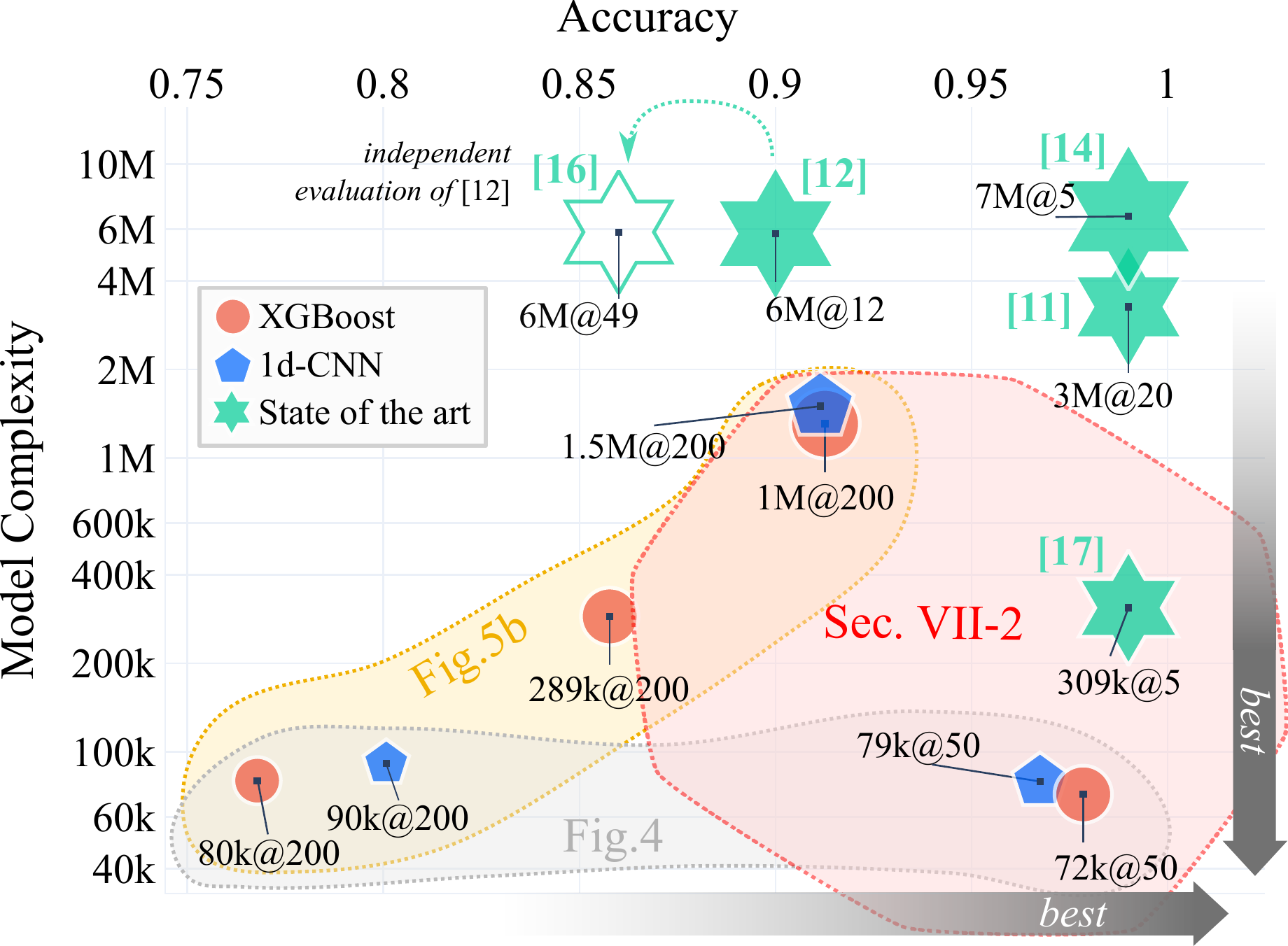}

\begin{minipage}{\columnwidth}
\centering
\scriptsize
\begin{tabular}{p{0.05cm} c r r r@{$\,\!$}l c}
\toprule
& 
\multicolumn{1}{c}{\bf Method} &
\multicolumn{1}{c}{$\boldsymbol{W}[\times1k]$}&
\multicolumn{1}{c}{$\boldsymbol{K}$} & 
\multicolumn{2}{c}{$\boldsymbol{W/K} [\times1k]$} &
\multicolumn{1}{c}{\bf Acc. [\%]}
\\
\midrule
\multirow{5}{*}{\rotatebox{90}{\begin{scriptsize}State of the art\end{scriptsize}}} &
\cite{chen2017bigdata}          
            & 6,640             & 5     & 1,328 &       & 99\\
&\cite{shapira2019infocom}
            & 309               & 5     & 61    &       & 99\\
&\cite{wang2017isi}          
            & 5,800             & 12    & 483   &       & 90\\
&\cite{wang2017icoin}
            & 3,270             & 20    & 163   &       & 99\\
&\cite{aceto2018tma}          
            & 5,800             & 49    & 119   &       & 86\\
\midrule
&\multirow{3}{*}{1d-CNN}        
             & 79                & 50    & 1     &.44    & 96\\
&            & 90                & 200   & 0     &.450   & 81\\
&            & 1,479             & 200   & 7     & .5    & 91\\
\midrule
&\multirow{4}{*}{XGBoost}    
             & 72                & 50    & 1     &.44    & 96\\
&            & 80                & 200   & 0     &.4     & 76\\
&            & 289               & 200   & 1     &.44    & 85\\
&            & 1,307             & 200   & 5     &       & 91\\
\bottomrule
\end{tabular}
\end{minipage}

    \caption{\emph{Models performance}. Scatter plot of model complexity vs accuracy,  for 1d-CNN (pentagons), and XGBoost  (circles). References from the State of the Art  (green stars) are included for \emph{qualitative} comparison purposes.
    Shapes are directly annotated ($W@K$) with the model size $W$ and classes $K$, and the shape size is proportional to the weights-per-class $W/K$ value (the table details the scatter plot values). 
    }
    \label{fig:complexity}

\end{figure}

The scatter plot in Fig.\ref{fig:complexity} illustrates the accuracy-vs-complexity trade-off for both our and literature models.  We underline that while the picture provides quantitative assessment of the accuracy and complexity performance, comparison among literature models should be interpreted  \emph{qualitatively} (as the datasets on which these results have been gathered differ). Results show that models in the literature are quite heavyweight, using up to \mbox{$W/K= 6.6\textrm{M}/15 = 1.3\textrm{M}$} and generally $W/K\gg100\textrm{k}$ weights-per-class, with the most parsimonious approach employing about 61k weights-per-class~\cite{shapira2019infocom}. In contrast, our CNN and XGBoost models can achieve about 90\% accuracy on the top-200 classes with at least two orders of magnitude less weights (just 5-7k weights-per-class), while still covering a $4\times$ bigger number of classes.

Two other observations emerge from the picture. On the one hand, when the number of classes is large ($K$$\approx$$200$), it becomes necessary to increase models size to maintain accuracy performance, but CNN and XGBoost models can be tuned to achieve similar performance (recall Fig.\ref{fig:ml-vs-dl}). 
On the other hand, when the number of classes is small ($K$$\approx$$50$), it is unreasonable to use an humongous number of weights to discriminate them (see Fig.\ref{fig:trainKtestX}),  particularly since it is possible to design parsimonious models with just hundreds of weights-per-class achieving same-or-better performance (i.e., bottom-right region of the scatter plot, which we further discuss in Sec.\ref{sec:discussion}). \emph{As such, by neglecting model complexity, the risk is to propose solutions that are  equivalent to ``shoot a mosquito with a cannon'', which may hinder deployability}.

\section{Detection of Unknown Applications}\label{sec:eval:nd}
\begin{figure}[t]
    \centering
    \includegraphics[width=0.99\columnwidth]{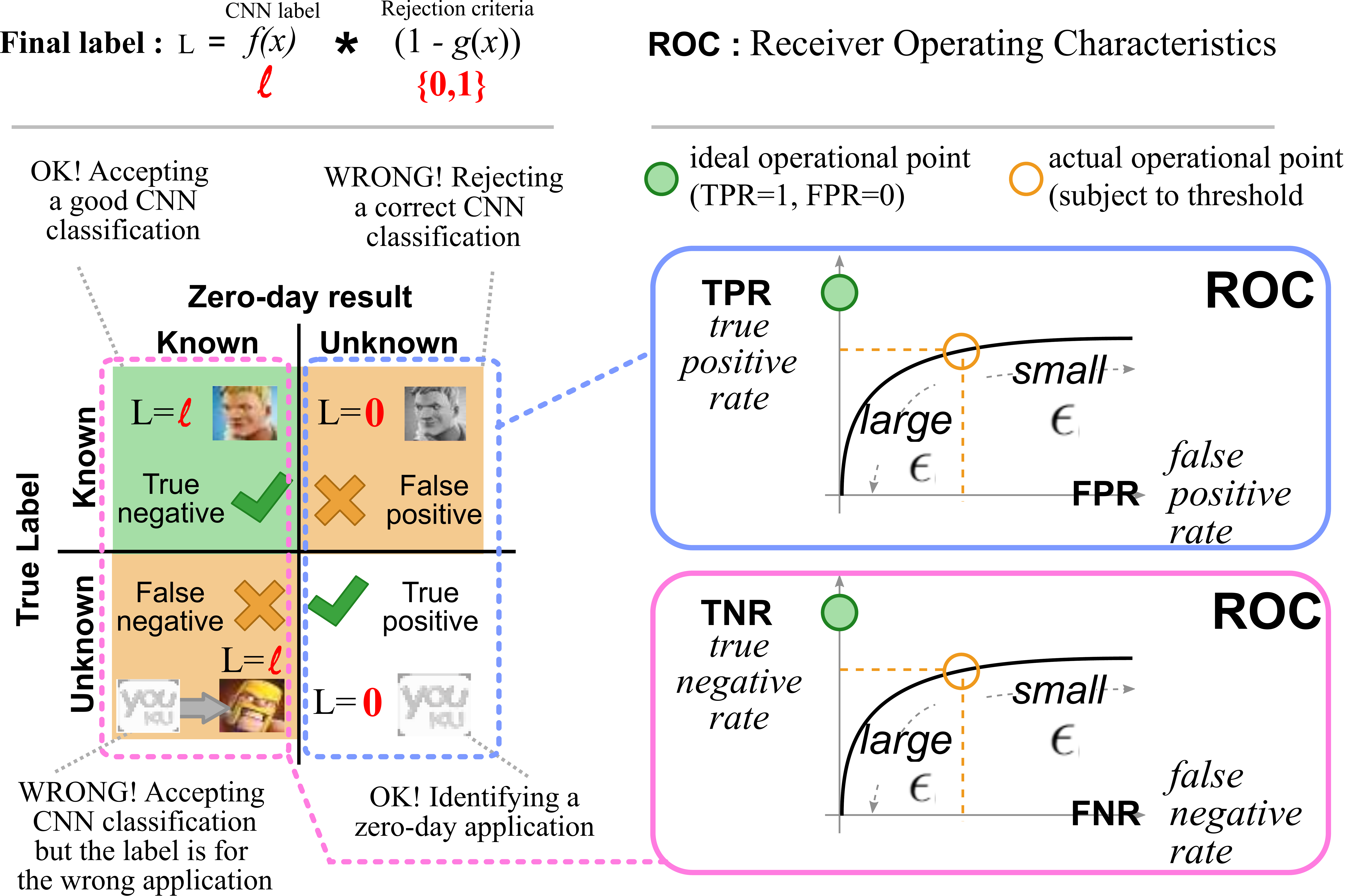}
    \caption{\emph{Quantifying zero-day application detection}: confusion matrix based on zero-day detection (left); ROC curve for zero-day application detection (right).}
    \label{fig:nd:sketch_tpr-fpr}
\end{figure}

\begin{figure*}[t]
    \centering
    \subfloat[UDP traffic]{\includegraphics[width=\columnwidth]{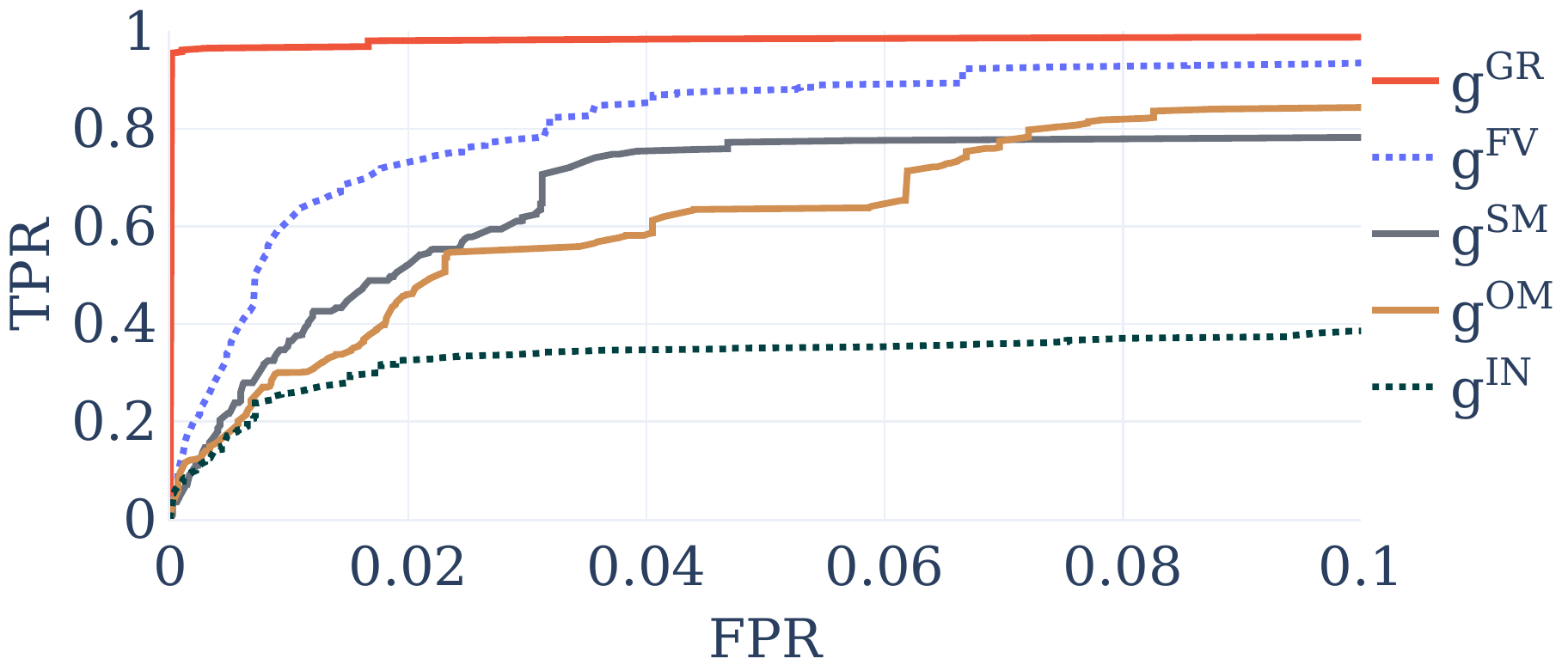}}~~ 
    \subfloat[TCP traffic]{\includegraphics[width=\columnwidth]{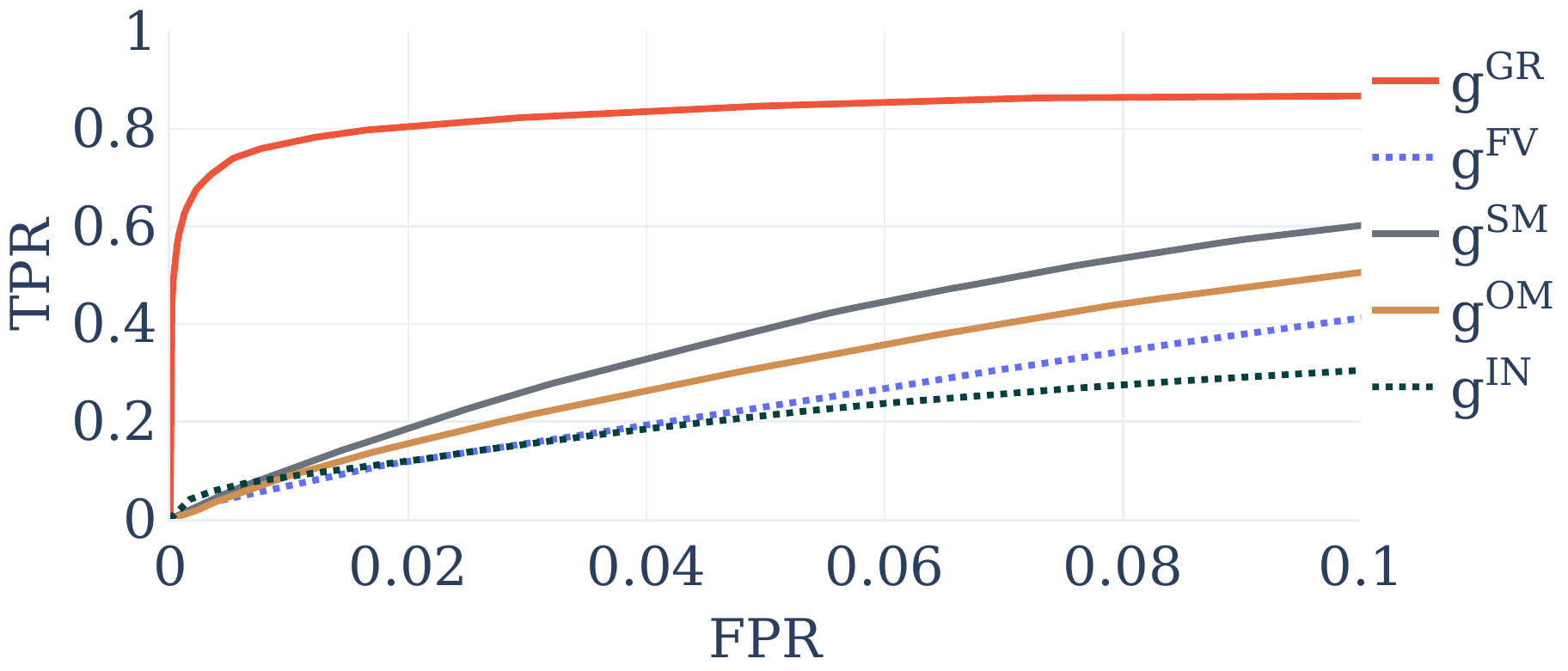}} \\
    \subfloat[UDP traffic]{\includegraphics[width=\columnwidth]{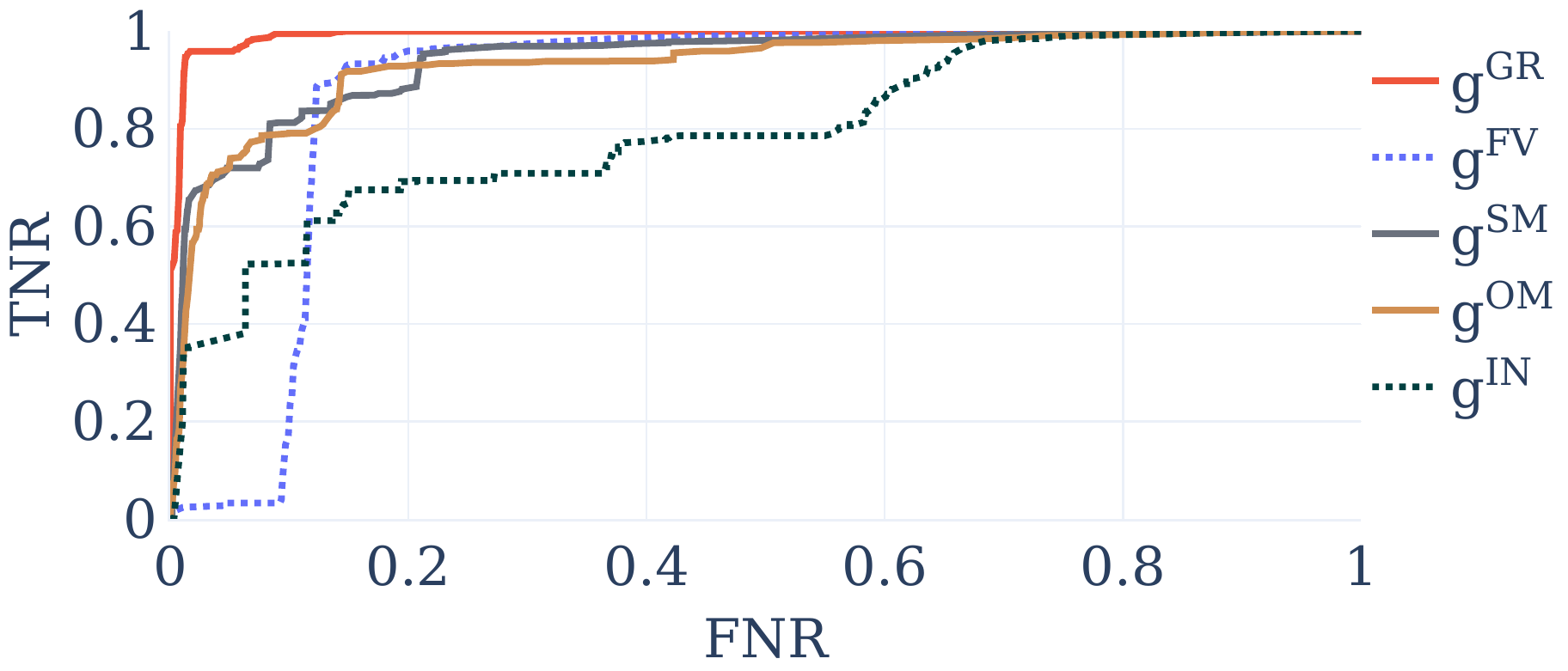}}~~ 
    \subfloat[TCP traffic]{\includegraphics[width=\columnwidth]{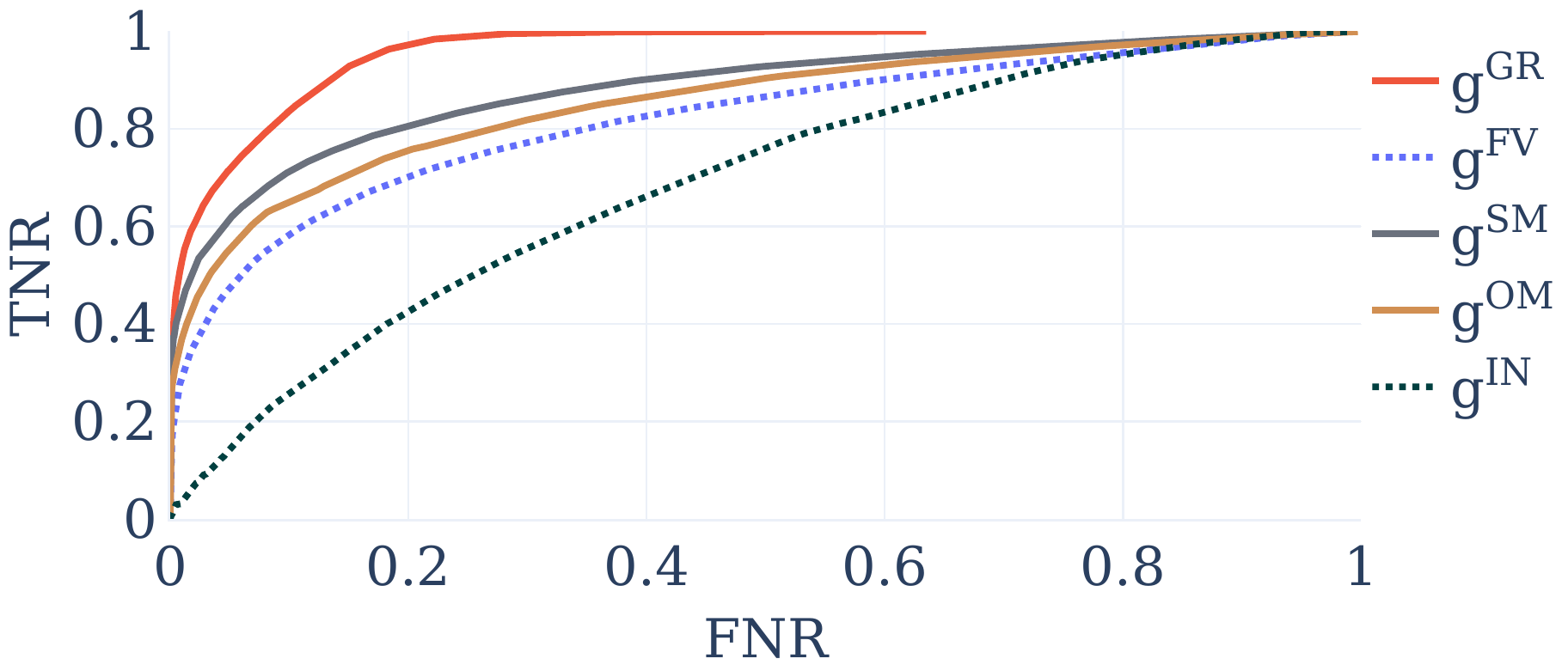}}

    \caption{\emph{Zero-day applications detection}: ROC curves  for UDP (left column) and TCP (right column) applications.  Two families of ROC curves are reported: TPR vs FPR (top row) and TNR vs FNR (bottom row).
    }
\label{fig:nd:roc}

\end{figure*}

In this section we evaluate zero-day applications detection. We built upon the CNN models evaluated in the previous section by applying the zero-day detection techniques introduced in Sec.\ref{sec:system:nd}: input clustering, feature vectors clustering, SoftMax, OpenMax, and our novel \OURS method. 
We first 
introduce the experimental workflow (Sec.\ref{sec:nd:workflow}) to overview zero-day detection across the methods  (Sec.\ref{sec:nd:highlevel}). Then we elaborate on the reasons behind their performance differences (Sec.\ref{sec:nd:deeper}), and we conclude evaluating their computational complexity (Sec.\ref{sec:nd:complexity}).

\subsection{Workflow}\label{sec:nd:workflow}
\subsubsection{Overall  comparison}
Fig.\ref{fig:nd:sketch_tpr-fpr} sketches our evaluation methodology. Given a trained model, we perform the classification of both known and unknown classes, and construct the Receiver Operating Characteristic (ROC) curve for all  novelty detection methods.
Usually,  ROC depicts the achievable True Positive Rate (TPR) on the y-axis, i.e., how well the model correctly identifies the 635 zero-day applications, for a given  False Positive Rate (FPR) on the x-axis, i.e., how often the model rejects a known top-200 application, wrongly relabeling it as zero-day application.  We additionally build a ROC curve to depict the complementary True Negative Rate (TNR) on the y-axis, i.e., the fraction of correctly accepted classification of known traffic, for a given False Negative Rate (FNR), i.e., the fraction of undetected zero-day traffic that are left mislabeled.

The ROC curves depict the different (FPR, TPR)  and (FNR, TNR) achievable operational points, by spanning over all possible values of the zero-day rejection thresholds $\epsilon$ discussed in Sec.~\ref{sec:system:nd}. Hence, the ROC curves are an intuitive way to compare different methods irrespective of the specific tuning: when the ROC curve for a method-$A$ is higher than the ROC for method-$B$, 
method-$A$ is better than method-$B$.

\subsubsection{Fixed tuning}
Any rejection method operates in fixed setting, i.e., it would be desirable to select a threshold $\epsilon$ enabling the system to operate as close as possible to 
the ideal (FPR,TPR) and (FNR, TNR) operational points (top-left corner of a ROC curve).
Therefore, we additionally compare methods at a \emph{fixed operational point} by selecting, for each method, the value of $\epsilon$ that corresponds to  a \emph{target  performance} in a  metric of interest. In detail, using the training dataset, we set the $\epsilon_X$ for method $g^{X}(\cdot)$ such that 99\% of classifications are accepted as TNR; respectively, tolerating 1\% of wrongly rejected FPR classifications. 

Whereas individual $\epsilon$ settings differ across methods, one can then relatively compare all  methods fairly by constraining their TPR when FPR=1\% (or their FNR when TNR=99\%). Importantly, as the above TNR=99\% and FPR=1\% targets only  depend on the known applications that the model has been trained with,  setting $\epsilon$ can be done without any prior knowledge of zero-day traffic.  

\subsection{High-level view}\label{sec:nd:highlevel}


\begin{figure*}[!t]
  \centering
  \subfloat[Input~space]{\includegraphics[width=.19\linewidth,valign=t]{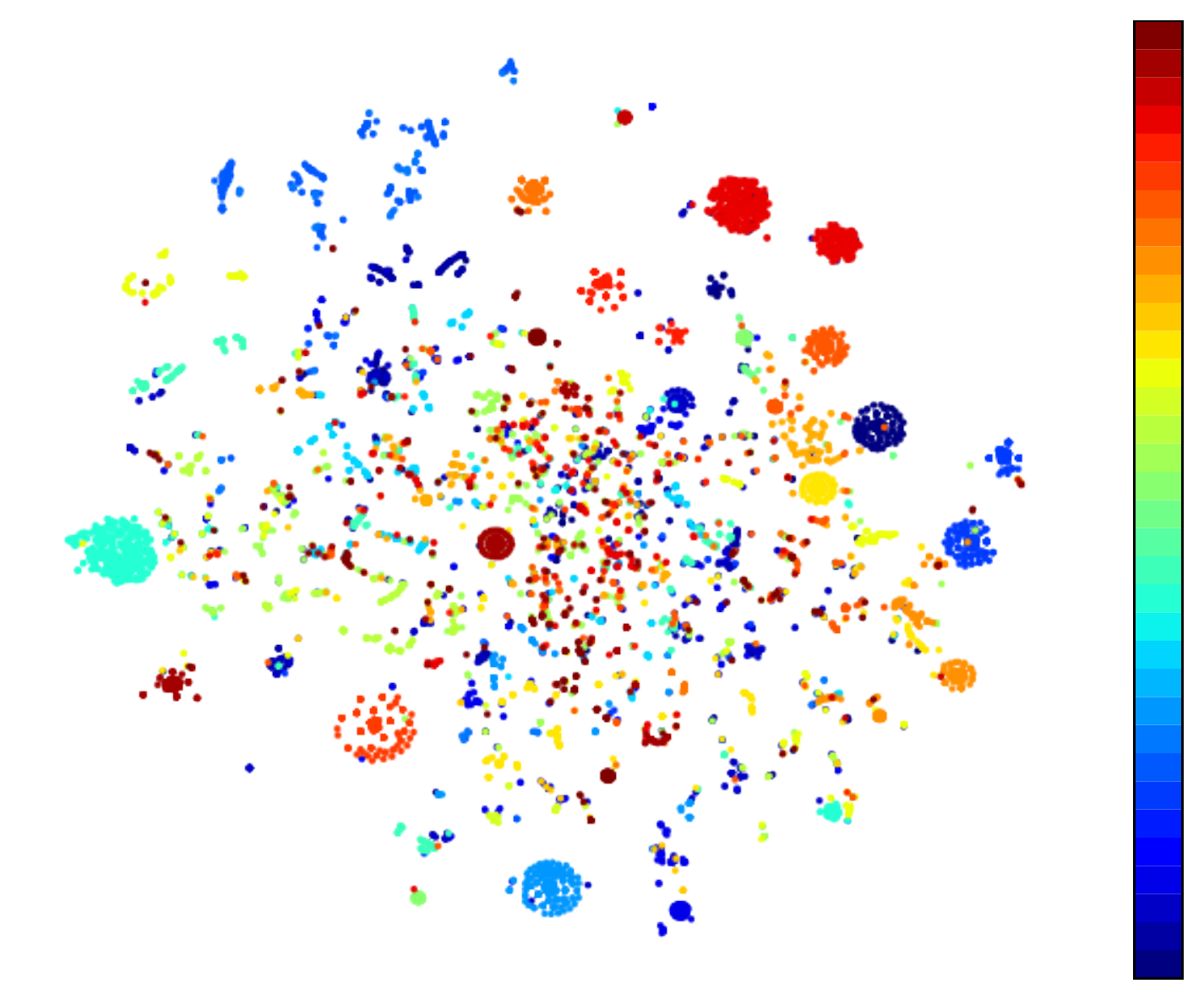}}
\subfloat[Feature Vector space]{\includegraphics[width=.19\linewidth,valign=t]{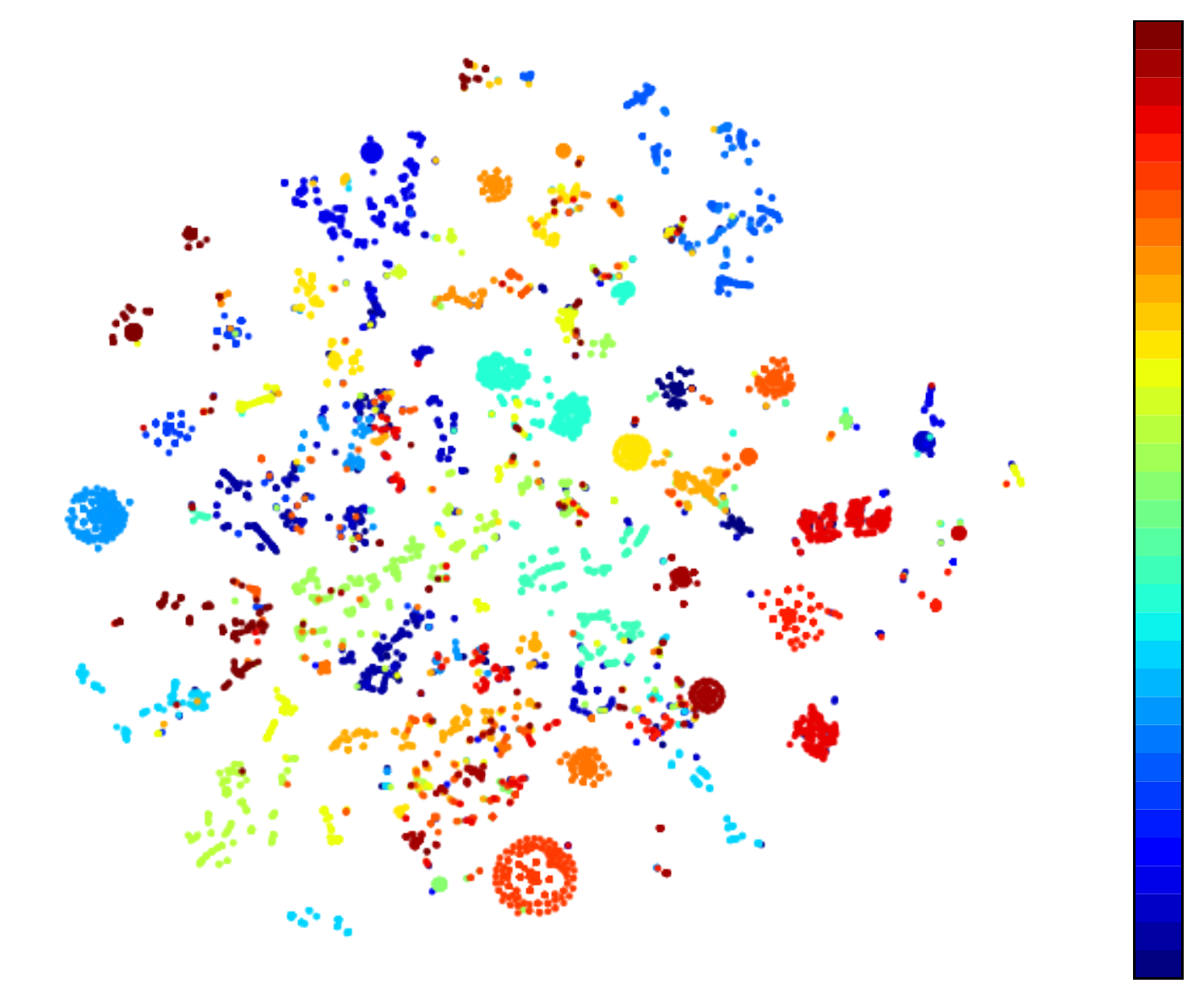}}
\subfloat[SoftMax]{\includegraphics[width=.22\linewidth,valign=t]{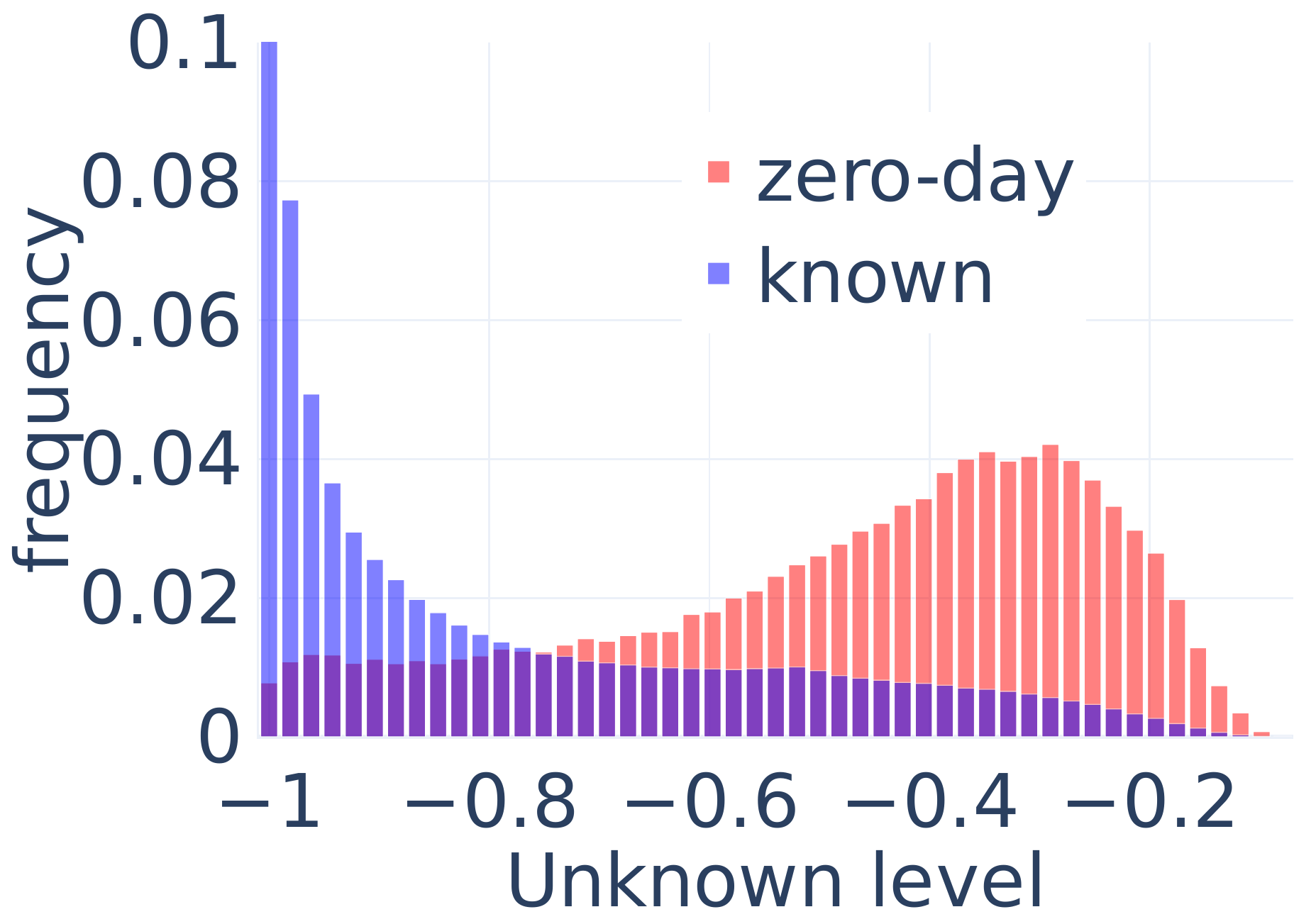}}
\subfloat[OpenMax]{\includegraphics[width=.22\linewidth,valign=t]{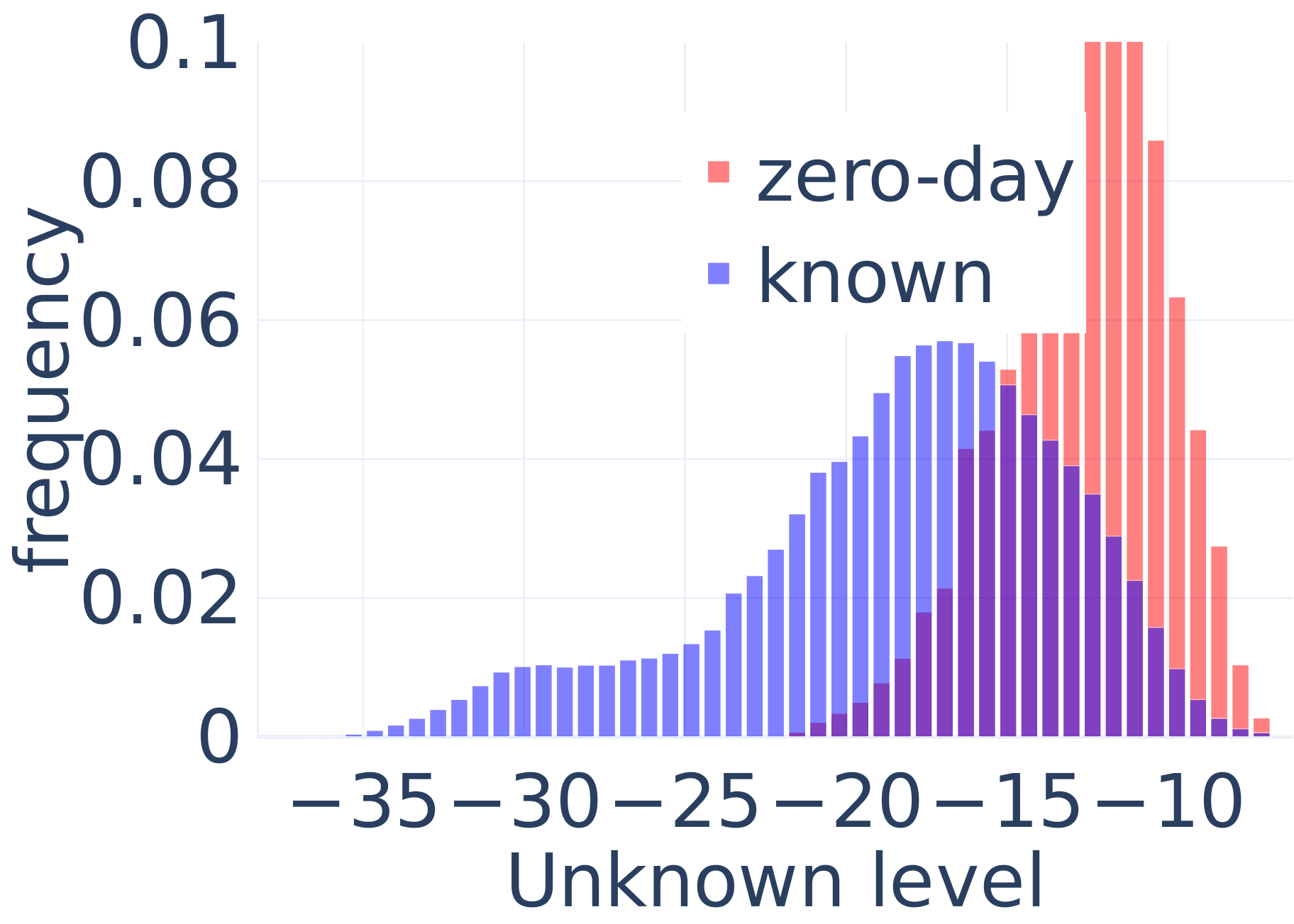}}
\subfloat[\OURS]{\includegraphics[width=.22\linewidth,valign=t]{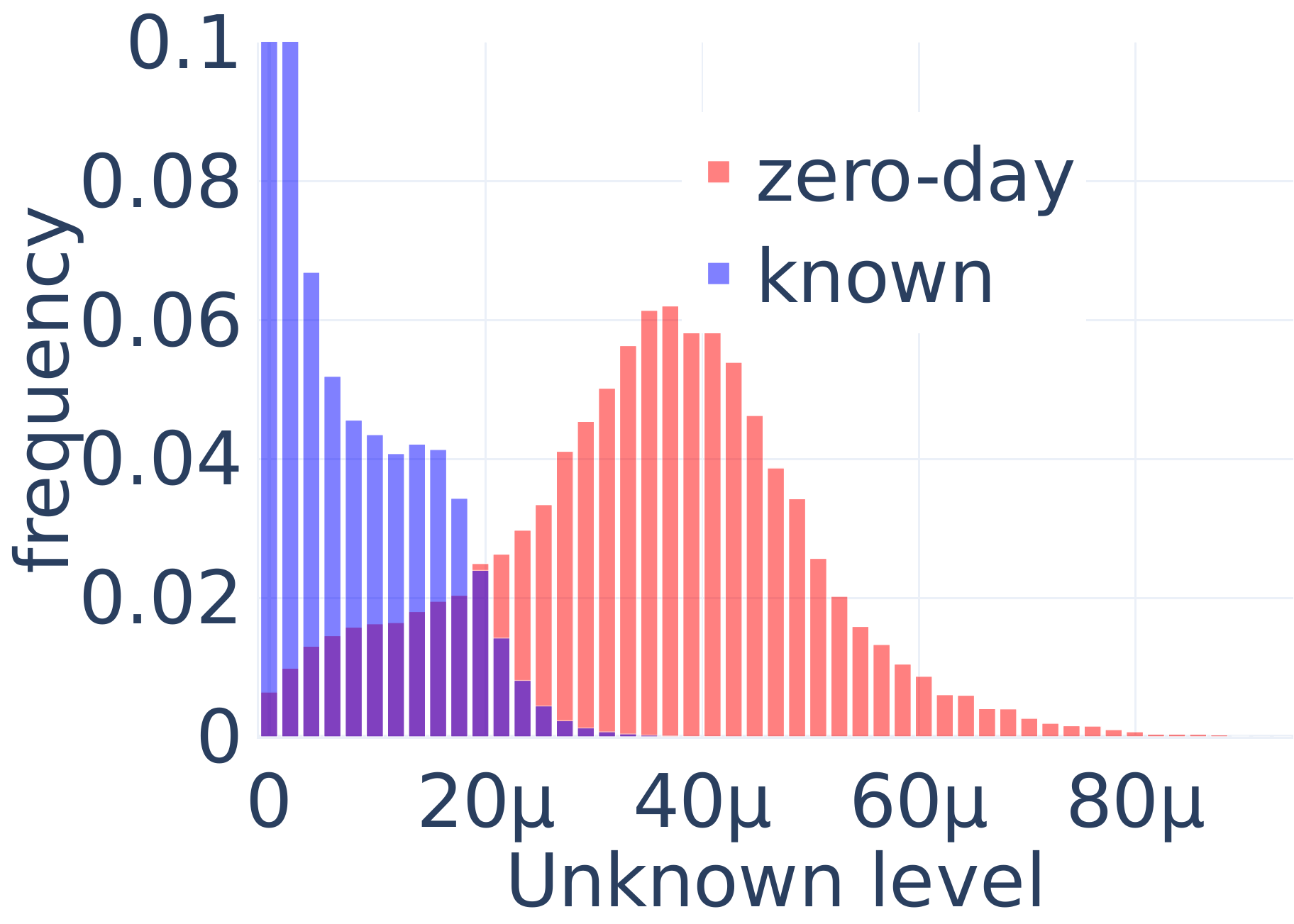}}
\begin{small}
\begin{tabular}{p{3cm}p{3cm}p{4cm}p{4cm}p{4cm}}
Purity:~0.91 &  Purity: 0.97 & Intersection: 0.39 & Intersection: 0.46 & Intersection: 0.21\\
Silhouette:~0.71 &  Silhouette: 0.72 & Bhattacharyya:  0.36 & Bhattacharyya:  0.31 & Bhattacharyya:  0.87 \\
\end{tabular}
\end{small}
\caption{\emph{Deeper insight on zero-day application detection}:  Qualitative visual differences of   (a) Input-clustering   and (b) FV-clustering via t-SNE projection (UDP application only), and histograms of unknown level for (c) SoftMax, (d) OpenMax and (e)  \OURS methods.  Relevant cluster (purity, silhouette) and distribution distance metrics (Intersection, Bhattacharyya distance) are used to quantitatively support the visual comparison.}
\label{fig:nd:hist}
\vspace{-2ex}
\end{figure*}

\begin{figure}[!t]
\centering
\includegraphics[width=\columnwidth]{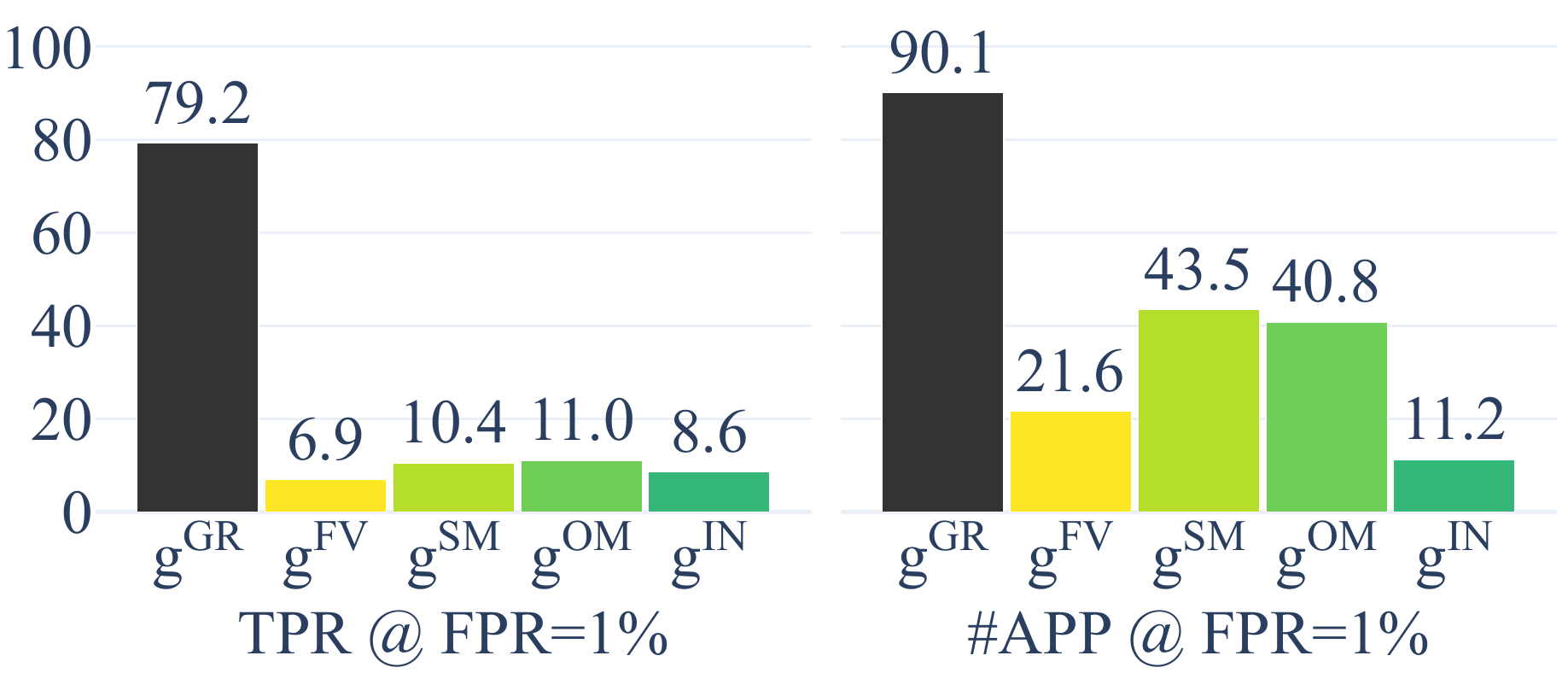}
\begin{tabular}{llllll}
\hline
$\epsilon_{GR}$ &   $\epsilon_{FV}$ & $\epsilon_{SM}$ & $\epsilon_{OM}$ &  $\epsilon_{IN}$ \\
$2.4\cdot10^{-5}$ & 14.9 & -0.23  & 0.34 & 4.9 \\
\hline
\end{tabular} 

    \caption{\emph{Zero-day applications detection}: Fraction TPR of correctly detected zero-day flows (left) and applications (right) when FPR=1\% of correctly identified known applications are rejected (the specific $\epsilon$  settings yielding the FPR=1\% target are tabulated in the picture for each method).
    }
    \label{fig:roc_grad}
\end{figure}

\subsubsection{Overall comparison} Recalling Table~\ref{tab:dataset:tcp-udp}, the TCP model can identify 162 known classes, and zero-day detection is tested against 162+500 known+unknown classes; the UDP model can identify 38 known classes, and zero-day detection is tested against 38+135 known+unknown  classes. Notice that input and feature vectors clustering require a specific training step to apply K-means: we use $C$=1,000 clusters (which is 5$\times$ the number of known applications, to account for multi-modal classes) and gather the cluster centroids from the same training data used to train the supervised model $f(x)$. 
The ROC curves depicting zero-day detection capabilities for all methods are reported in Fig.~\ref{fig:nd:roc}, for both UDP and TCP, and for both types of ROC curves (notice that in the FRP-vs-TPR ROC, we visually limit FPR$<$10\% as higher values would represent unreasonable operational points).

It is immediate to gather that \OURS ($g^{GR}$) outclasses all alternatives by a large margin: \OURS performance is almost perfect for UDP (note that TPR=96\% already when FPR=0.1\%) and remains good in the more challenging TCP scenario (TPR=78\%, FPR=1\%) especially compared to alternative approaches.  We tuned OpenMax as recommended by the original authors~\cite{Bendale15}, i.e., with a tail size of 20 when fitting the $\omega_b$ Weibull distribution. Yet, differently from~\cite{Bendale15}, SoftMax outperforms (or marginally differs from) OpenMax.
Clustering the input data ($g^{IN}$) results is the worst option. Instead, the performance for clustering the feature vectors ($g^{FV}$) varies depending on the scenario: for UDP, it is the second best option; for TCP, it does not appear to be better than alternatives. 
While increasing the number of clusters may provide benefits\cite{Zhang15}, as we shall see, the current setting already has a prohibitive computational cost (Sec.~\ref{sec:nd:complexity}), which makes increasing the number of clusters hardly viable.


\subsubsection{Fixed tuning} 
Fig.\ref{fig:roc_grad} compares the different methods performance when focusing on the \mbox{top-200} applications (i.e., combining the TCP and UDP models), and by settings $\epsilon$  so to obtain FPR=1\%. The left-side of the figure measures zero-day detection per-flow; the right-side measures zero-day classes detection, where a zero-day class is considered detected if at least one of its flows is detected. While \OURS correctly detects 79.2\% of all zero-day flows (90.1\% of the classes) the other approaches are only able to detect at most 11\% zero-day flows (43.5\% of classes).  

Additionally, Fig.\ref{fig:nd:roc} and Fig.\ref{fig:roc_grad} also suggest that tolerating a higher FPR would only  have marginal improvements and that \OURS is less sensible to threshold tuning. 
In fact, while for all methods the TPR varies depending on the FPR, for \OURS the TPR increases very quickly for small FPR, and grows slowly afterward, i.e., after the curve ``knee'' it is almost insensitive to the configured threshold.




\subsection{Deeper insights}\label{sec:nd:deeper}

Results discussed so far 
clearly show differences between  methodologies, but the underlining reasons causing the performance gaps are not obvious. We therefore dig into the specific  details of each method, starting from clustering. Recalling  Fig.\ref{fig:nd:roc}, input-clustering is the worst performing method. This is expected since the non linear transformations operated by the CNN feature extraction indeed aim to empower better separation in the latent space than what could be achieved in the input space. 

We can however visualize this effect by mean of a bi-dimensional  t-SNE~\cite{maaten2008tsne}. Fig.\ref{fig:nd:hist}-(a,b) t-SNE plots show the 38 UDP known applications, each associated to a different color. Both plots show sharp clusters, but notice how the center of the input space (Fig.\ref{fig:nd:hist}-a) is more confused than the FV space (Fig.\ref{fig:nd:hist}-b).
We can further quantify this effect using \emph{clusters purity} (the average fraction of the dominant class in each cluster) and \emph{clusters silhouette} --- the metric captures how similar a class is to its own cluster (\emph{cohesion}), compared to other clusters (\emph{separation}). We find that the latent space has the highest purity (0.97-vs-0.91), but presents a similar silhouette with respect to the input space (0.71-vs-0.72). 


To investigate the three remaining methods we measure the \emph{unknown level} which quantifies predictions uncertainty: following~(\ref{eq:softmax}), the unknown level for SoftMax is the negative of the maximum probability $UL^{SM} = -max_{k\in[1,K]}P(y=k|x)$; following (\ref{eq:openmax}), the unknown level for OpenMax is the synthetic activation $UL^{OM} =\hat{v}_0(x)$; finally, following~(\ref{eq:gradient}), for \OURS  the unknown level is the magnitude of the backpropagation update $UL^{GR} = \|\delta^L\|_n$. 

Fig.~\ref{fig:nd:hist}-(c,d,e) show the 
distribution of the unknown level for known and unknown classes.
To further quantify the overlap of the distributions, we annotate the figure with the \emph{Intersection index} $I$ 
\begin{equation}
    I = \sum \min(UL_{\mathrm known}, UL_{\mathrm zero-day})
\end{equation}

\noindent and the  \emph{Bhattacharyya distance} $D$
\begin{equation}
    D = -ln \sum \sqrt{UL_{\mathrm known} * UL_{\mathrm zero-day}}
\end{equation}


Ideally, $g(x)$ should be able to well separate the unknown level distributions for known and zero-day classes, or equivalently $g(x)$ should have a low intersection index $I$, and high distance $D$. Intuitively, known classes should provide a small unknown level. For instance, the unknown level for SoftMax is capped at -1 which expresses a strong confidence in the model prediction. Conversely, when processing traffic of a zero-day application, the confidence is expected to reduce. The same is true for OpenMax and Gradient-based, although the unknown levels are unbounded.  We can see that distributions overlap for OpenMax, leading to an higher $I$,  and a lower $D$ than for the other methods, which explains the lower performance.
This large confusion hints that the activation vectors  might have a large diversity for the known classes, hence a simple mean activation vector~(\ref{eq:openmax}) is insufficient. Similarly, \OURS presents half the $I$ of SoftMax, while $D$ is over $2\times$ larger, which explains the better performance of \OURS.


    
\subsection{Complexity}\label{sec:nd:complexity}
\begin{table}[!t]
\centering
    \caption{\emph{Zero-day application detection complexity}: \\
    $g(x)$  bootstrap and inference costs }
    \begin{tabular}{llrc} 
\toprule
\multirow{2}{*}{\bf Method} & 
\multirow{2}{*}{\bf g(x)}   & 
\bf Bootstrap cost & 
\bf Inference cost  \\
&  
& 
\bf (per dataset) & 
\bf (per sample) \\
\midrule    
\multirow{2}{*}{Gradient}  
                    & $g^{GR}_{u,2}$    & -       & 0.01~ms\\
                    & $g^{GR}_{u,1}$    & -       & 0.01~ms\\
\midrule    
SoftMax             & $g^{SM}$          & -       & 0.01~ms\\
OpenMax             & $g^{OM}$          & 11s     & 0.13~ms\\
FV-clustering       & $g^{FV}$          & 24m~36s & 4.58~ms\\
Input-clustering    & $g^{IN}$          & 20m~4s  & 4.46~ms\\
\bottomrule    
    \end{tabular}
    \label{tab:nd:complexity}
\end{table}

We conclude comparing  initialization costs and inference computational costs as summarized in Table~\ref{tab:nd:complexity}. Results are obtained using TensorFlow v1.9 on an NVIDIA P100 GPU. SoftMax and \OURS do not need any specific pre-processing, OpenMax introduces a small cost to fit the Weibull distribution, and clustering-based methods (as expected) have the largest initialization cost (recall that those use $K$-means with $C$=1,000 clusters in our setup). 

We report initialization costs just for reference. In fact, given their episodic and offline nature, they are less relevant from a practical perspective than inference computational costs. In particular, we find
 OpenMax being $13\times$ slower than \OURS, and clustering-based methods being more than two orders of magnitude slower than our method. Indeed, they suffer from the computation of a large number (specifically $C$) of pair-wise distances (involving power of two, and square root operations) in a fairly large space (for TCP, the FV space has 128 dimensions, and input space has 100 dimensions).
We point out that while distance computations can be reduced (e.g., smaller latent space, int8 quantization, Manhattan distance, etc.) the performance of clustering-based methods can be hardly expected to improve. In fact, while increasing the number of clusters $C$ might improve accuracy performance, this would increase their computational cost too. 

\OURS therefore brings the best of both worlds, as it is not only significantly more accurate, but also computationally lightweight.
Specifically, for a TCP+UDP model with $K$=200 classes, \OURS performs an Hadamard product between vectors of size 200 in (\ref{eq:gradient}), and a matrix multiplication between a 128$\times$200 and 200$\times$1 matrices, followed by a Hadamard product between vectors of size 128 in (\ref{eq:gradient2}). Additionally, while the computational costs results are to be interpreted in a relative sense (i.e., across algorithms), we stress that the raw  performance  reported here comes from a non optimized implementation --- yet,  using TensorFlow v1.9 on a NVIDIA P100 GPU, we reached 100,000 $g(x)$ inferences per-second which is enough to sustain real system requirements~\cite{gallo2020sigcomm}.

\section{Summary and Discussion}\label{sec:discussion}
 In this paper, we focused on the creation of a  commercial-grade traffic classification engine capable of ($i$) fine-grained applications identification of hundreds of classes and ($ii$)  zero-day detection of applications that were not part of models knowledge base. 
 We tested the engine on a large dataset using the 
 top-200 applications (covering 95\% of the flows and bytes) for the identification of known classes,
 and top-835 applications (extending the flows and bytes coverage to over 99\%) for zero-day detection.
 
Summarizing our main findings, we gathered that ($i$)  ML and  DL  models  are  both  equally  able  to  provide  satisfactory solutions for the classification of known traffic. In particular, the type of CNN architectures used in the literature are well suited as their accuracy exceeds 90\% for the top-200
applications. At the same time, our results also pointed out that models complexity is commonly overlooked in literature. This yields to models that are  unnecessarily complex for relatively simple tasks, endangering the practical relevance of the research.

More interesting, we gathered that ($ii$)  ML and DL differ in their abilities to detect zero-day traffic, as  the non-linear  feature  extraction  process  of  DL models backbone yields to sizable  advantages over ML for this task. In particular, our main contribution was to provide a novel gradient-based technique for zero-day detection tailored for DL classifiers, that we found to outperform the related literature in both accuracy and computational costs.

Despite our efforts, we still identify a few key open questions, related to ($i$) open datasets, ($ii$) models complexity, ($iii$)  novelty discovery techniques and ($iv$) models deployability, that needs further discussion.

\fakepar{Open dataset --- pooling efforts} 
In reason of our results, constructing an open corpus with rich \emph{class diversity} (and thus necessarily  a large \emph{class cardinally $K$}) should be a priority goal to allow for meaningful and fair cross-comparison of research proposals.
While this is within reach for large industrial players, legal and business aspects prevent them to share openly their datasets.

For instance, we are currently investigating the possibility to release 
a highly anonymized version of the dataset used in this paper,
as  part of  Huawei Rapid Analytics \& Model Prototyping (RAMP)
Data Challenges.\footnote{These data challenges, hosted at \url{http://xianti.fr}, 
are a new Huawei initiative, to exactly facilitate this type of data sharing
through the quite common format of data science challenges popularized by Kaggle \url{https://kaggle.com}.}
However, while the dataset we used does not contains information that constitutes a 
\emph{privacy-related} risk (e.g., IP addresses, timestamps), it contains
\emph{business-sensitive} information  (e.g., fine grained labels and packet size sequences). Such information requires further processing (e.g., label obfuscation, time series shuffling, etc.), to 
ensure that  researchers can still carry meaningful experimental activities 
(e.g., use the dataset as benchmark for their neural architectures), while not endangering the business at the same time (e.g., if  relevant labels were open-sourced, this would allow to train and openly release models of commercial value).

This said, we observe that while collecting large volumes of real labeled network data is a daunting effort for a \emph{single academic} partner, a pooling effort across \emph{multiple research groups $N$} in a coordinated manner can be an effective strategy to achieve this goal --- for instance, each partner can  gather $K/N$ classes, coordinating to keep a null/low class overlap between groups $\mathcal{K}_i\cap \mathcal{K}_j = \emptyset$. Also, different research groups  are already doing active measurement collections for specific application types (video, games, etc.) and with a different goal than traffic classification (congestion control, QoE, etc.), so that the true burden lies in the coordination. This is commonplace in other communities (e.g., ImageNet has 15 million images labeled in 20k classes), and the traffic classification community should take inspiration from those efforts.  This would allow to tackle more challenging classification problems, with a more significant application diversity and a set of fine-grained labels.

\fakepar{Complexity --- the need for models ecology} 
For the classification of known traffic, in this paper we decided not to participate to the arm race against literature (as we admittedly did in the first wave), and we avoided to propose yet another ``new'' model for traffic classification. Quite the opposite~\cite{troublingtrends}, we decided to perform the simplest and  meaningful selection of relevant DL and ML models (based on our own experience, as well as on consolidated results  reported in the state of the art). This choice allowed us to clearly state a very simple yet important message ---
as far as classification of known traffic is concerned,  ML and DL models are equally well fitted to support large-scale commercial scenarios.

Given our choice to limit the set of considered models, in this paper we intentionally avoided answering the question of which is the model that best fit traffic classification. We argue that while most of the research in DL focused on 
models \emph{accuracy} comparison (recall Table~\ref{tab:related:tc}), we believe that it is now necessary to explore more systematically the \emph{accuracy}-vs-\emph{complexity} trade-off. In particular, we observe that the ``Green networking''~\cite{greenNet2010} research wave 
(i.e., the explicit consideration of energy expenditures in network protocols, algorithms, and architecture designs) predates by over a decade the corresponding ``Green AI'' wave~\cite{greenAI2020}.
To fully understand models complexity we need to go beyond the analysis reported in this paper, and in  literature in general. In particular, as early pointed out in Sec.\ref{sec:eval:tc}, models size does not directly translate into neither models execution time (as operations differ across models) nor models energy expenditure (which depends on the hardware).
Therefore, we argue that further research is needed to relate models accuracy with their computational complexity --- this requires to  additionally consider metrics such as \emph{classification/Joule} or \emph{accuracy/Watt}.

At the same time, we identify many open research questions regarding the \emph{design} of more efficient DL architectures (bi-dimensional convolutions~\cite{shapira2019infocom},
sparse-LSTM~\cite{Hua2019}, 
point-wise convolution as in ShuffleNet\cite{ma2018shufflenetv2},
inverted residual as in MobileNet\cite{howard2018mobilenetv2}, etc.) and on how to improve their \emph{implementation efficiency} considering traffic processing challenges (as in \cite{jouppi2017sigarch,tensorflow-lite-micro} for the general case, and  \cite{gallo2020sigcomm,gallo2021sec} for traffic classification in particular). As exploring this space is time consuming, automated Neural Architecture Search (NAS)~\cite{elsken2019nas} techniques result appealing. However, NAS is guided by an indirect measure of computation complexity (i.e., FLOPs). Therefore,  further research is needed to explicitly include in the NAS loop more direct metrics,  such as speed or energy consumption, to perform \emph{ecology} of models design space.  


\fakepar{Novelty discovery --- the broader picture} 
The main contribution of this paper is represented by \OURS, a novel DL-tailored technique for zero-day traffic detection. We found that, while DL and ML are equally good for the classification of known traffic, DL-gradient based techniques may give to DL methods a significant advantage over ML methods for the detection of unknown traffic.

We also believe that \OURS opens an interesting avenue for further research.
For instance, in our initial design, we experimented by  backpropagating to more than one layer. In particular, we found that running a \emph{full backpropagation} renders GradBP  significantly slower (about $4\times$ in our 1-d CNN models)  and it additionally reduces accuracy (by almost 10\%). While the extra computational cost is somewhat expected, the accuracy drop is more intriguing, and 
we believe it relates to the known \emph{vanishing gradient} problem: the more the layers traversed with backpropagation, the smaller the gradients, the greater the difficulty to measure differences. 

Whereas \OURS experimental results show that focusing on the last layer allows to catch large modifications and suffices for zero-day detection of known applications,
we argue that the design space for alternatives is still open.  For instance, \emph{weighting layers} might reduce the vanishing gradient problem. However, more research is needed to devise simple and effective algorithms to precisely state how layers should be weighted. Additionally, the potential accuracy gains should be contrasted with the extra computational complexity (recall that zero-day detection is  additional work to accumulate to the detection of known classes, thus should be minimal). 

Finally, while in this paper we only applied our proposed  technique for the purpose of zero-day detection, we argue  that its application is more general.
As such, it would be interesting to evaluate \OURS in the scope of open-set recognition, e.g., by casting its application into computer vision where DL models have initially flourished.


\fakepar{Deployability --- the elephant in the room} 
Given the maturity of traffic classification research,  
 focusing only on raw classification performance, albeit of novel DL models, 
 does not help academic models to step out of academic venues. Instead, it would be imperative to tackle other pressing problems that impact models deployability in the real world~\cite{pacheco18comst}. 
 
 A key aspect of deployability is the \emph{model training loop}. For instance, 
 recalling Fig.\ref{fig:system},
 it is clear that known applications will evolve and eventually some will become unpopular. Similarly, new (zero-day) applications  will emerge, for which only a few samples will be initially available. In ML/DL terms, this maps to problems where learning \emph{evolves over time}, such as  in \emph{continuous learning}  (e.g., to tackle the knowledge drift of existing classes~\cite{carela2016streaming})  or in \emph{incremental/decremental learning} (e.g., to  add zero-day applications~\cite{tma2021} or to remove ``old'' classes from an existing model).
 Due to the heavy-tailed nature of the applications popularity, samples of zero-day applications will be scarcer than for known classes. This calls for experimentation with \emph{few-shot learning}~\cite{fewshot} techniques  to extend models beyond popular classes.
 
 Additionally, deployability raises problems where learning needs to be \emph{distributed over space}, as in  \emph{federated learning} (e.g., for privacy or business-sensitive constraints~\cite{yang2020ijcai,fedavg}, or due to applications heterogeneity~\cite{yang2020ijcai} and multi-modality as shown in Sec.\ref{sec:dataset:multimodal}).

 Finally, beside models complexity, deployability also relates to \emph{model inference latency}, especially critical to support line rate analytics, and 
   the \emph{auditing of classification decisions} for the network experts (unlikely experts in the ML/DL domain too). In fact, DL models are known to be harder to interpret~\cite{beliard2020infocom,meng2020sigcomm}, which directly translates to the need for \emph{models explainability}~\cite{xaisurvey}, an aspect not covered in this work.
 
 In our experience, the ability to seamlessly and automatically incorporate new applications into a model, providing as much insights as possible to domain experts, is paramount for the successful transfer of research into products and deployments.

\def\UrlBreaks{\do\/\do-}

\section*{Acknowledgement}
We wish to thank the Editor and the anonymous Reviewers, whose feedback improved the quality of this paper.

\bibliographystyle{IEEEtran}
\bibliography{tc+nd-biblio}

\balance

\begin{IEEEbiography}[{\includegraphics[width=1in,height=1.25in,clip]{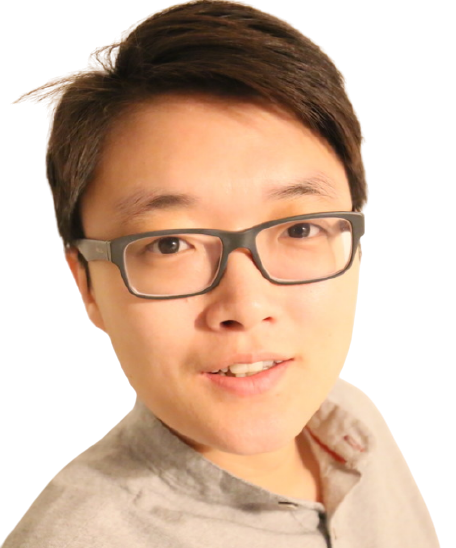}}]{Lixuan Yang}
is a Senior Engineer in the DataCom Lab at Huawei Technologies, France. She received her Ph.D. from the CNAM (2017) and MSc from Jean Monet University(2013). She worked at the XXII Group as an artificial Intelligence team leader on computer vision and natural language processing projects. Her current interests include traffic classification, federated learning, continual learning.
\end{IEEEbiography}

\begin{IEEEbiography}[{\includegraphics[width=1in,height=1.25in,clip]{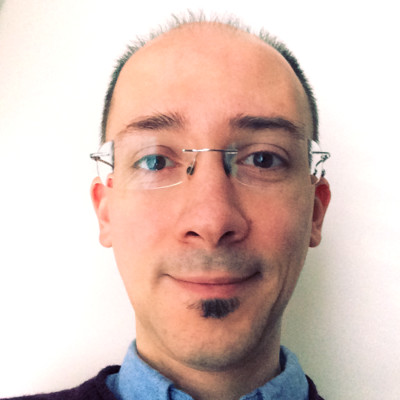}}]{Alessandro Finamore}
is a Principal engineer in the DataCom Lab at Huawei Technologies, France. Before joining Huawei in 2019, he was a Principal engineer at Telefonica UK/O2 (London, United Kingdom), and an associate research at Telefonica Research (Barcelona, Spain). He received his Ph.D from Politecnico di Torino (2012). He has coauthored more than 80 papers in leading conferences and journals, including a best paper award at CoNEXT.
\end{IEEEbiography}

\begin{IEEEbiography}[{\includegraphics[width=1in,height=1.25in,clip]{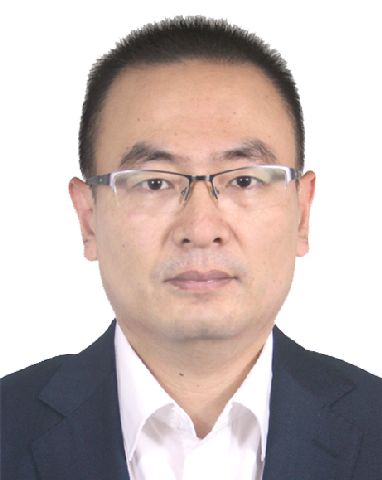}}]{Jun Feng}
Feng Jun is a data scientist and AI architect expert in Huawei Data Communication Product Line. He has over 20 years of experience in data analysis and AI deployment in the telecommunications industry.
\end{IEEEbiography}

\begin{IEEEbiography}[{\includegraphics[width=1in,height=1.25in,clip]{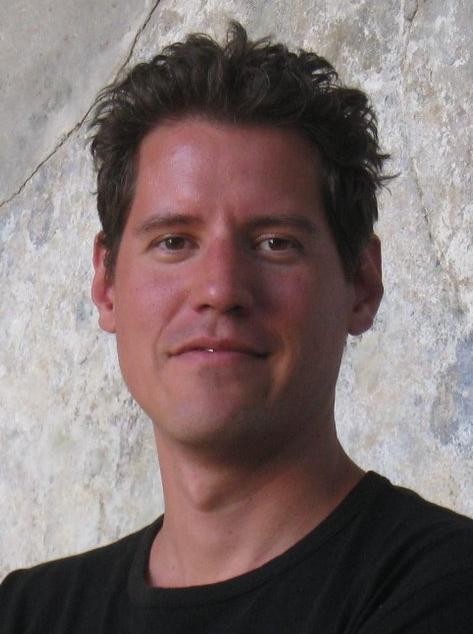}}]{Dario Rossi}
is Network AI CTO and Director of the DataCom Lab at Huawei Technologies, France. Before joining Huawei in 2018, he held Full Professor positions at Telecom Paris and Ecole Polytechnique and was holder of Cisco's Chair NewNet\@ Paris. He has coauthored 15 patents and over 200 papers in leading conferences and journals, that received 9 best paper awards, a Google Faculty Research Award (2015) and an IRTF Applied Network Research Prize (2016). He is a Senior Member of IEEE and ACM.
\end{IEEEbiography}

\end{document}